\documentclass[logo,msc]{infthesis}  % <-- adjust degree tag if needed
\usepackage{msccheck}   % keep page-limit guard
\usepackage{microtype}  % recommended
\usepackage{booktabs}
\usepackage{tabularx}
\usepackage{float}
\usepackage[section]{placeins}
\usepackage{longtable}
\usepackage[final]{graphicx}
\usepackage[T1]{fontenc}
\usepackage[utf8]{inputenc}
\renewcommand{\arraystretch}{1.2}
\usepackage{tikz}
\usetikzlibrary{positioning,fit,arrows.meta,shadows}
\usepackage{url}
\usepackage[hidelinks]{hyperref}

\begin{document}
\begin{preliminary}

\title{Benchmarking LLM Agents in Wealth-Management Workflows}

\author{Rory Milsom}

\date{\today}

\abstract{

Modern work relies on an assortment of digital collaboration tools, yet routine processes continue to suffer from human error and delay. To address this gap, this dissertation extends TheAgentCompany with a finance-focused environment and investigates whether a general-purpose LLM agent can complete representative wealth-management tasks both accurately and economically. This study introduces synthetic domain data, enriches colleague simulations, and prototypes an automatic task-generation pipeline. The study aims to create and assess an evaluation set that can meaningfully measure an agent's fitness for assistant-level wealth management work. We construct a benchmark of 12 task-pairs for wealth management assistants spanning retrieval, analysis, and synthesis/communication, with explicit acceptance criteria and deterministic graders. We seeded a set of new finance-specific data and introduced a high- vs. low-autonomy variant of every task. The paper concluded that agents are limited less by mathematical reasoning and more so by end-to-end workflow reliability, and meaningfully affected by autonomy level, and that incorrect evaluation of models have hindered benchmarking.}

\maketitle

\newenvironment{ethics}
   {\begin{frontenv}{Research Ethics Approval}{\LARGE}}
   {\end{frontenv}\newpage}

\begin{ethics}
% <<< Keep one of the following two blocks; delete the other. >>>

%---- If ethics approval WAS required ----
% This project obtained approval from the Informatics Research Ethics committee.\\
% Ethics application number: ???\\
% Date when approval was obtained: YYYY-MM-DD\\
% Participants' information sheet and consent form are in the appendix.

%---- If ethics approval was NOT required ----
This project was planned in accordance with the Informatics Research Ethics policy.
It did not involve any aspects that required approval from the Informatics Research
Ethics committee.

\standarddeclaration
\end{ethics}

\begin{acknowledgements}

I thank Dr Alexandra Birch-Mayne and Barry Haddow for their great guidance, and the maintainers of TheAgentCompany, OpenHands, OwnCloud, Rocket.Chat, Plane, and EspoCRM for making this work possible. I am especially grateful to my family, particularly my Mum and Dad for their constant encouragement and support. Finally, I am incredibly grateful for my beautiful girlfriend for putting up with me for this stressful year, as well as her unwavering belief in me, even when I didn't believe in myself.

\end{acknowledgements}

\tableofcontents
\end{preliminary}

%==================== MAIN CONTENT (≤ 40 pp) ====================

\chapter{Introduction}\label{chap:Introduction}

\section{Motivation}\label{sec:motivation}
Financial institutions are actively exploring autonomous, tool-using language models to support everyday operations; compiling reports, reconciling client data, drafting communications, and coordinating meetings \cite{mckinsey2023genaiBanking}. The appeal is clear: these workflows are frequent, time-sensitive, and document-heavy; automation promises lower latency - leading to quicker results, reduced cost, and consistent audit trails. At the same time, finance is a safety-critical and regulated industry. Errors in data handling, calculations, or communications can create real financial and compliance risk. This tension (obvious potential utility versus high stakes) explains \emph{why this matters}: before adoption, we need credible evidence that such systems reliably complete realistic work to spec.

An AI agent is a Large Language Model (LLM) augmented with tools (browser, file I/O, APIs), short-term memory, and a decision policy that iterates a plan–act–observe loop. Unlike a pure chat model, an agent can complete complex workflows, for example, fetch CSVs from a document store, query a CRM, compute totals in Python, and send a confirmation message, end-to-end execution rather than isolated text generation. In finance settings the agent’s \emph{relevance} is precisely this ability to operate across systems that mirror real back-office workflows (document store, CRM, chat, ticketing).

Given the stakes, we argue that rigorous, reproducible evaluation must precede deployment. Ad-hoc demos can be misleading; a wealth-management focused benchmark is necessary that fixes the tools, data, and acceptance criteria so we can measure reliability, efficiency, and failure modes in a controlled way. Prior to this, TheAgentCompany demonstrates how to run agents inside a simulated workplace \cite{xu2024theagentcompany}, and OpenHands provides a widely used agent runtime \cite{openhands2024}. We build on that idea but orient it squarely toward finance: we create an evaluation set that reflects assistant-level wealth-management tasks and grade them deterministically.

We focus on assistant tasks because they admit objective checks, for example, computing a net-worth snapshot, reconciling portfolio weights, and preparing year-end summaries from source ledgers, all of these common tasks have verifiable outcomes, which is important for evaluation. These jobs require cross-tool retrieval, basic analysis, and precise delivery, yet they can be verified by concrete criteria (file exists at the specified path with the required schema; numbers match recomputation; message is sent in the required format). That property makes them suitable for automated, reproducible assessment.

In summary, our goal is not to deploy an agent but to \emph{construct and validate an evaluation set} that financial practitioners could trust: realistic domain-aligned tasks, clear inputs and outputs, and deterministic graders inside a stable, tool-rich environment. This provides the evidence base needed to judge whether present-day agents meaningfully help with wealth-management assistant work, and where they still fall short. \emph{Preview of findings:} three patterns recur in our results—(i) end-to-end workflow reliability (finding files, respecting paths, delivering outputs) limits performance more than arithmetic; (ii) reducing autonomy via schema/path specificity curbs drift without inflating search depth; and (iii) strict-but-tolerant graders (schema/format exactness with numeric/CSV normalisation) avoid both false positives and brittle byte-match failures. Limitations include a closed-world setup with synthetic but normalised data, a small client/task set, and primary evaluation on a single agent configuration.

\section{Research Objective}\label{sec:research-objective}
Our primary objective is to create and assess an evaluation set that can meaningfully measure an LLM agent’s fitness for assistant-level wealth-management work. Concretely, we ask:

\begin{quote}
\textbf{Can we create a challenging task suite to effectively test an LLM’s ability to perform as a financial employee}
\end{quote}

A secondary question examines prompt design: \emph{What is the effect of agent autonomy on LLM accuracy and efficiency in a simulated workplace?} To study this, we run otherwise identical agents in a controlled, tool-rich testbed and compare checkpoint pass-rates, completions, steps, and cost across paired versions of the same tasks. Throughout, we treat a \emph{checkpoint} as a grader-verifiable assertion and a task as \emph{resolved} only when all checkpoints pass. We execute the suite inside a seeded, open-source, closed-world stack (EspoCRM~\cite{espocrm}, OwnCloud~\cite{owncloud}, Rocket.Chat~\cite{rocketchat}, and Plane~\cite{plane}).

\section{Contributions \& Outcomes}\label{contributions-outcomes}
\begin{itemize}
  \item \textbf{Domain-aligned evaluation set.} We construct a benchmark of 12 task-pairs for wealth-management assistants spanning retrieval, analysis, and synthesis/communication, with explicit acceptance criteria and deterministic graders. The suite is designed to reflect real assistant workflows, for example, including EspoCRM ~\cite{espocrm}, while remaining reproducible and auditable.
  \item \textbf{Reproducible agent environment.} We adapt an existing agent-in-environment framework to a finance context \cite{xu2024theagentcompany,openhands2024}, seeding realistic data and fixing service identities and paths (OwnCloud~\cite{owncloud}, Rocket.Chat~\cite{rocketchat}, Plane~\cite{plane}, and including EspoCRM ~\cite{espocrm} - for a more realistic finance work environment) so that runs are comparable over time and across models. Outcomes reflect improved task suite design and agent competence, not easier tasks.
  \item \textbf{Autonomy study.} We compare high- vs.\ low-autonomy versions of the \emph{same} tasks. Low-autonomy briefs generally improve checkpoint accuracy and completions with modest changes in steps and cost, indicating that clearer specifications help translate reasoning into deliverables; persistent failures highlight remaining system bottlenecks.
\end{itemize}

\section{Structure of the Document}\label{sec:structure-of-document}
Chapter~\ref{chap:background} sets the context with related work on agentic LLMs, evaluation in regulated domains, and agent-in-environment benchmarks. Chapter~\ref{chap:methodology} describes the evaluation set: task design, data seeding, grader construction, and the controlled environment used for experiments. Chapter~\ref{chap:results-&-evaluation} reports results for the legacy-vs-new comparison and the autonomy study, including quantitative metrics and error analysis. Chapter~\ref{chap:conclusion} summarises findings, limitations, and directions for future work on harder workflows, stronger verification, and broader external validity.

\chapter{Background and Related Work}\label{chap:background}

\paragraph{Overview.}
This chapter (i) introduces what we mean by LLM \emph{agents} and \emph{autonomy}, (ii) motivates the shift from abstract, single-shot evaluations to realistic workplace settings, and (iii) describes \emph{TheAgentCompany} (TAC) in depth, including its environment, task design, checkpoint evaluation, and headline results, before turning to finance-specific benchmarks and cost–accuracy trade-offs.

\section{Agents and Autonomy}\label{sec:agents-autonomy}

We use \emph{agent} to mean an LLM-centred system that runs an observe–reason–act loop: it reads observations (UI state, files, chat), plans using the model, and executes actions (browser clicks, shell commands, Python code, or messages) without a human scripting each step \cite{russell2020aima,wang2024llm_agent_survey}. \emph{Autonomy} is meant as the extent to which that loop can take a task from intent to completion with minimal intervention, or brevity of initial prompt (forcing agents to have more.

In practice, modern LLM agents implement this via intertwined reasoning and tool use (the “ReAct”-style pattern), invoking tools or UIs as part of their chain of thought and adjusting future steps based on feedback \cite{yao2023react,schick2023toolformer,openhands2024}. Self-monitoring or reflection mechanisms are often added so agents can detect errors and course-correct mid-run \cite{shinn2023reflexion}.

In enterprise contexts, autonomy matters because real work is long-horizon and tool-mediated: the agent must sequence steps across multiple applications, carry context over time, and recover from errors; success depends on planning and interaction, not just text prediction \cite{wang2024llm_agent_survey,wang2023voyager}. Chat-only QA benchmarks therefore overestimate readiness for workplace use, they test language competence, not end-to-end task completion under realistic friction.

Evaluation therefore needs verifiable \emph{checkpoints} that capture intermediate progress (e.g. “found the file”, “messaged the stakeholder”) and expose failure modes (auth errors, wrong endpoints, brittle UI navigation). Benchmarks that combine tool use, long-horizon plans, and collaboration give a truer picture of utility and safety \cite{yao2024taubench,xu2024theagentcompany}.

\section{Workplace Agent Benchmarks}\label{sec:workplace-agent-bench}
We focus on the field’s move \emph{from abstract/scientific single-shot tasks to testing usefulness in realistic work environments}. Recent research has begun evaluating large language model (LLM) agents on realistic workplace tasks \cite{wang2024llm_agent_survey}. A notable example is TheAgentCompany (TAC), a benchmark that simulates a modern digital workplace environment \cite{xu2024theagentcompany}. Xu et al. construct a self-contained “small software company” intranet populated with typical tools e.g. GitLab (code repo), OwnCloud (documents), Rocket.Chat (team chat), and Plane (project management), all seeded with realistic data to mirror a multifaceted office setting \cite{gitlab,owncloud,rocketchat,plane}. Within this environment they defined 175 diverse tasks spanning roles like software engineering, project management, human resources, finance, and administration \cite{xu2024theagentcompany}. Tasks range from coding and data analysis to form-filling and inter-office communication, thereby testing agents’ broad capabilities in a professional context \cite{xu2024theagentcompany}. Crucially, TheAgentCompany uses the open-source OpenHands agent framework \cite{openhands2024}, enabling agents to browse web pages, run code in a Jupyter kernel, execute shell commands, and interact with simulated co-worker agents via chat \cite{kluyver2016jupyter}. This allows evaluation of not only tool use but also collaboration – agents can ask LLM-based “colleagues” for help or information through the chat interface, reflecting the social dynamics of real workplaces \cite{xu2024theagentcompany}. To handle the complex, long-horizon tasks, the benchmark employs a checkpoint-based evaluation: each task is broken into verifiable subtasks with deterministic or LLM-based evaluators, awarding partial credit for intermediate progress \cite{xu2024theagentcompany,yao2024taubench}. This design captures realistic multi-step workflows and provides nuanced performance scores beyond all-or-nothing outcomes.

\begin{figure}[H]
  \centering
  \makebox[\textwidth][c]{%
    \includegraphics[height=0.22\textheight,keepaspectratio]{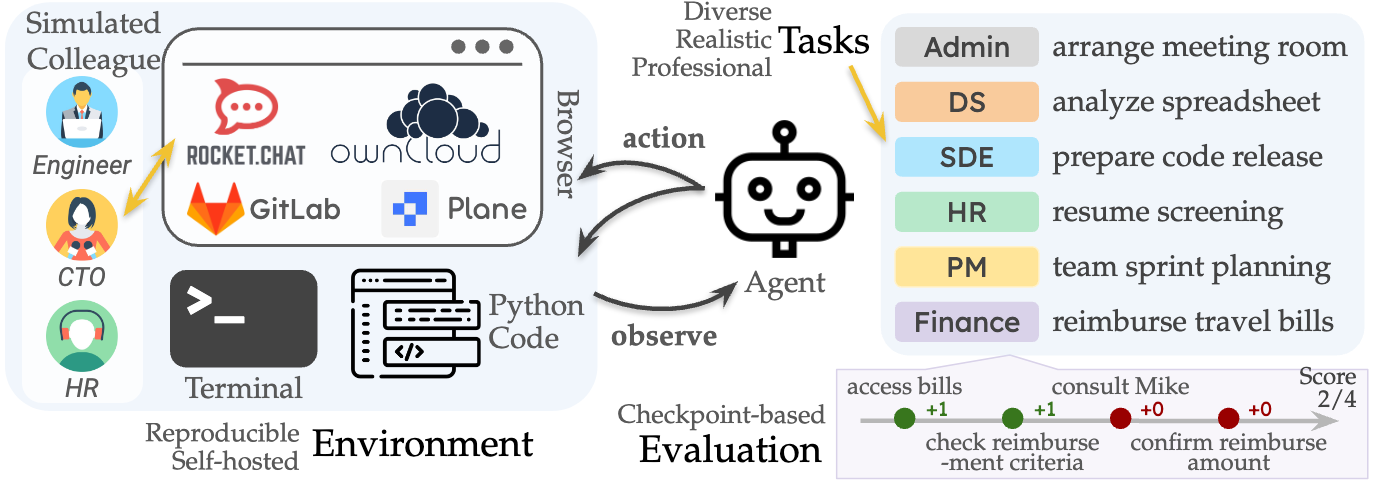}%
  }
  \caption{Overview of TAC Architecture\cite{xu2024theagentcompany}}
  \label{fig:tac-architecture}
\end{figure}

\noindent\textbf{Deeper TAC description.}
Beyond the high-level summary, TAC fixes an identical agent scaffold (OpenHands) and swaps only the LLM backend, ensuring fair comparison across models. The intranet is reproducible and self-contained, so results are not confounded by live-web variance. Each task includes: (i) a natural-language brief, (ii) seeded artefacts (documents, repos, tickets, chats), (iii) role-realistic actions (e.g. updating a Plane issue, sending a Rocket.Chat DM, filling a PDF in OwnCloud), and (iv) a checklist of verifiable checkpoints culminating in a completion verdict. TAC reports both \emph{full} completion and \emph{partial} scores from checkpoint credit. Using seven different LLMs as backends (including GPT-4, Gemini, and Llama), the best agent (Claude 3.5) fully solves only 24\% of tasks and reaches about 34\% with partial credit, with the steepest drop on long-horizon and business/finance/admin tasks \cite{xu2024theagentcompany,openai2023gpt4,team2023gemini,dubey2024llama3,anthropic2024claude35}. These results indicate substantial headroom before agents can reliably handle non-trivial office workflows.

Beyond TheAgentCompany, several other benchmarks have explored LLM agents in workplace-like settings, each with a different focus. WorkArena evaluates web-based agents on common office tasks using live browser interactions, though it does not incorporate multi-step task structure or inter-agent communication \cite{drouin2024workarena}. SWE-Bench and DevBench target software engineering tasks in coding environments \cite{jimenez2023swebench,li2024devbench}, while AssistantBench and Yao et al.’s tool-use benchmark focus on dialog and tool-augmented agents in customer service domains \cite{yoran2024assistantbench,yao2024taubench}. These efforts demonstrate growing interest in “enterprise agent” evaluation, but each has limitations. Many prior benchmarks cover a narrow scope of skills or domains (e.g. only coding or only web form tasks) and often lack the long-horizon, checkpointed structure and collaborative aspect that TheAgentCompany introduced \cite{xu2024theagentcompany,yao2024taubench}. In contrast, TheAgentCompany combines a diverse task spectrum with realistic workplace interactions, setting a higher bar for agent evaluation in professional contexts. However, it primarily centres on a software company scenario; thus, important domains like finance receive only cursory coverage. This opens the opportunity to extend such benchmarks into domain-specific settings, as our work does for the financial sector.

\section{Financial Domain Agent Evaluation}\label{sec:financial-domain}
While general agent benchmarks are emerging, financial tasks remain under-represented in prior evaluations. There is a growing recognition of this gap, and a few recent works have begun to address it. For instance, Xie et al. introduce FinBen, a comprehensive benchmark aggregating 42 datasets across 24 financial NLP tasks (e.g. information extraction, question answering, forecasting) \cite{finben2024}. FinBen’s broad evaluation of 21 models (GPT-4, ChatGPT, Google Gemini, etc.) revealed that current LLMs excel at basic financial text processing but struggle with advanced reasoning tasks like long-form report generation or market forecasting \cite{finben2024}. Notably, even top models perform unevenly,  GPT-4 was strong in structured tasks like information extraction and even stock trading simulations, yet it faltered on complex decision-making, highlighting the difficulty of expert financial reasoning \cite{finben2024}. FinBen underscores the need for domain-specific benchmarks, but it primarily consists of standalone dataset evaluations rather than an integrated agent environment.

Researchers have also started developing agent-oriented financial benchmarks. Bigeard et al. (2025) propose a Finance Agent Benchmark that poses realistic financial research problems to tool-using LLM agents \cite{bigeard2025fab}. They crafted 537 expert-written queries (e.g. analyzing recent SEC filings) across a taxonomy of nine finance task categories \cite{bigeard2025fab}. Agents are equipped with tools like web search and direct access to the SEC EDGAR database to simulate a financial analyst’s workflow \cite{bigeard2025fab}. This benchmark revealed significant limitations: the best model reached only $\sim$46.8\% accuracy, and achieving even this modest performance required expensive API usage (averaging \$3.79 per query in model fees) \cite{bigeard2025fab}. Another contemporaneous effort is InvestorBench (Li et al., 2024), which evaluates LLM-based agents on financial decision-making tasks (e.g. stock or cryptocurrency trading decisions) in various market scenarios \cite{li2024investorbench}. It provides a suite of environments and datasets for investment-focused reasoning and tests multiple LLMs’ abilities to adapt across different financial products \cite{li2024investorbench}. These domain-specific benchmarks mark important first steps toward assessing AI agents in finance. They demonstrate that, much like in general workplaces, current models fall short on complex finance tasks – often requiring human-level expertise, multi-step quantitative reasoning, or domain-specific knowledge that LLMs lack. Moreover, the high inference costs reported (e.g. several dollars per query for near state-of-the-art agents) raise concern for practical deployment in finance, where efficiency is paramount \cite{bigeard2025fab}.

Despite these advances, there remains a clear gap in benchmarking LLM agents as financial assistants performing everyday professional duties. Existing finance benchmarks either focus on narrow decision problems (e.g. trading simulations) or evaluate models on isolated question-answer pairs. They do not fully capture the workflow of a finance office role – which might involve pulling data from spreadsheets, generating client reports, updating accounting ledgers, communicating findings, and so on in a continuous, multi-step process. Our work addresses this gap by creating a finance-oriented benchmark built on TheAgentCompany’s paradigm \cite{xu2024theagentcompany}. We design realistic tasks grounded in the daily responsibilities of financial analysts or advisors (for example, preparing year-end tax summaries, analyzing investment portfolios, or auditing expense reports). These tasks are embedded in an interactive environment with relevant data sources (financial statements, transaction ledgers, etc.), requiring the agent to perform tool-assisted calculations and document preparation akin to a real financial assistant. By evaluating agents on this suite of finance-specific challenges, we extend prior workplace benchmarks into a critical domain that was previously underexplored. This allows us to measure not only raw accuracy on finance problems, but also an agent’s ability to navigate the nuanced processes and tools unique to finance workflows – providing a richer assessment of utility in financial settings.

\section{Cost–Accuracy Trade-offs in Agent Evaluation}\label{sec:cost-accuracy}
Kapoor et al.\ argue that benchmarks should score both accuracy \emph{and} cost; otherwise agents can chase tiny gains with unbounded API calls \cite{kapoor2024agentsmatter}. They advocate evaluating models on a cost–accuracy Pareto frontier and show that simple policy tweaks can cut calls substantially with little loss, so leaderboards should favour cost-effective agents rather than accuracy at any price \cite{kapoor2024agentsmatter}.

\noindent Our evaluation follows this: we log per-task cost (tokens/tool invocations) alongside success, place models on a cost–accuracy curve, and surface trade-offs (e.g. larger models may win more tasks but at higher expense) \cite{openai2023gpt4}. This is especially important in finance, where time and budgets are tight; tracking cost lets us prefer agents that “do more with less,” in line with \emph{AI Agents That Matter} \cite{kapoor2024agentsmatter}.

\chapter{Methodology \& System Design}\label{chap:methodology}  % 

\section{Methodology Overview}\label{sec:methodology-overview}
The goal of this methodology is to measure whether LLM agents can complete realistic wealth-management assistant workflows in a reproducible, tool-rich environment. Achieving this requires: (i) a controllable agent runtime, (ii) seeded line-of-business services (CRM, document store, chat, ticketing), (iii) a task suite with explicit inputs/outputs and checkpointed evaluators, and (iv) an automated harness to launch runs and grade outcomes. To minimise confounds, we build directly on TheAgentCompany (TAC) and its OpenHands-based stack, then extend it for the finance domain.

\noindent\textbf{What we inherit from TAC (used largely as-is).}
\begin{itemize}
  \item \emph{Agent runtime:} OpenHands agent container with terminal, Python/Jupyter, browser, and file tools wired into a plan–act loop.
  \item \emph{Docker orchestration:} a top-level \texttt{docker-compose.yml} that boots auxiliary services (Rocket.Chat, Plane, OwnCloud, etc.) with named volumes.
  \item \emph{Task skeleton:} per-task directories containing \texttt{instruction/task.md}, a checkpointed evaluator (Python), and build artefacts (\texttt{Dockerfile}, \texttt{dependencies.yml}, \texttt{Makefile}).
  \item \emph{Service seeding:} per-service \texttt{Dockerfile}/\texttt{init.sh} and helper scripts (e.g., seeding personas from \texttt{npc\_definition.json}); \texttt{servers/populate\_data.py} inserts those NPCs into the backing store used by Rocket.Chat so IDs/channels are stable across runs.
  \item \emph{Evaluation helpers:} configuration (\texttt{config.toml}) and runner scripts (\texttt{run\_eval.py}, browsing helpers, \texttt{summarise\_results.py}) that start the agent, check service health, and emit per-checkpoint results.
\end{itemize}

\noindent\emph{Workspaces utilities (summary).} \texttt{workspaces/common.py} provides minimal helpers for Rocket.Chat (DM/channel history checks), OwnCloud (WebDAV list/check/download), Plane (basic project/issue lookups), a lightweight LLM predicate check, and a \texttt{@grader} wrapper. \texttt{workspaces/config.py} centralises endpoints, credentials, and a \texttt{TEST\_MODE} flag (credentials for NPC users are taken from TAC’s provided files). \texttt{workspaces/eval.py} is the evaluator entrypoint: it decrypts and loads the task-specific \texttt{grade\_checkpoints()}, optionally reads the trajectory, and writes \texttt{result.json}. \texttt{workspaces/scoring.py} defines \texttt{Checkpoint}/\texttt{Result} and a few simple scoring strategies. On the services side, \texttt{servers/setup.sh} performs a minimal bootstrap (preflight Docker checks, pull images, start the API server, and wait for Rocket.Chat/OwnCloud/Plane health).

\noindent\textbf{What we add/change (finance-focused extension).}
\begin{itemize}
  \item \emph{Domain realism:} introduce a live EspoCRM service seeded with client records/attachments; expand OwnCloud under \texttt{/Finance\_Documents} with ledgers and templates relevant to wealth management; provision additional Rocket.Chat colleagues (via Sotopia) and a small set of Plane issues to drive cross-tool workflows.
  \item \emph{Task suite:} a 24-task benchmark—12 \emph{high-autonomy} (brief) and 12 \emph{low-autonomy} (schema/path-explicit) variants—tiered across three difficulty levels (retrieval, analysis, synthesis/communication).
  \item \emph{Grading fidelity:} deterministic, checkpointed evaluators that verify file locations, schema conformance, numeric correctness, and (where applicable) chat/issue actions; step and cost accounting are recorded for each run.
  \item \emph{Reproducibility:} seeded named volumes and fixed IDs/paths across services; minor harness tweaks (e.g., pre-flight checks, preserving chat logs) to ensure bit-for-bit reruns and consistent scoring; specifically, the default Rocket.Chat reset in \texttt{npc/init.sh} is disabled so message history persists for audit.
\end{itemize}

This chapter then expands each element in turn: the base TAC/OpenHands stack we rely on, the finance extensions (services and data), the paired task design and difficulty tiers, the checkpoint evaluators, and the execution/analysis pipeline used in the experiments.

\section{Environment for Development}\label{sec:environment}

\begingroup
\setlength{\intextsep}{6pt}
\setlength{\textfloatsep}{6pt}

\begin{figure}[!htb]
\centering
\resizebox{0.72\linewidth}{!}{%
\begin{tikzpicture}[
  font=\small,
  node distance=20mm and 25mm,
  service/.style={draw, rounded corners=5pt, minimum width=3cm, minimum height=13mm, align=center, fill=white, thick, drop shadow={shadow xshift=1pt, shadow yshift=-1pt, opacity=0.3}},
  runtime/.style={draw, rounded corners=5pt, minimum width=14cm, minimum height=18mm, align=center, fill=blue!12, very thick, drop shadow={shadow xshift=1pt, shadow yshift=-1pt, opacity=0.3}},
  monitor/.style={draw, rounded corners=5pt, minimum width=3.5cm, minimum height=13mm, align=center, fill=green!20, thick, drop shadow={shadow xshift=1pt, shadow yshift=-1pt, opacity=0.3}},
  agent/.style={draw, rounded corners=5pt, minimum width=3cm, minimum height=13mm, align=center, fill=orange!20, thick, drop shadow={shadow xshift=1pt, shadow yshift=-1pt, opacity=0.3}},
  >={Stealth[length=3mm, width=2.5mm]}
]

% --- nodes/edges unchanged ---
\node[service] (own) at (0,0) {\textbf{OwnCloud}\\File Storage};
\node[service] (crm) at (3.5,0) {\textbf{EspoCRM}\\Customer Data};
\node[service] (chat) at (7,0) {\textbf{Rocket.Chat}\\Messaging};
\node[service] (plane) at (10.5,0) {\textbf{Plane}\\Project Mgmt};

\node[agent, minimum width=2.3cm, minimum height=10mm] (sotopia) at (7,1.8) {\footnotesize\textbf{Sotopia}\\{\tiny AI Agents}};

\node[runtime] (runtime) at (5.25,-3.5) {
  \Large\textbf{OpenHands Agent Runtime}\\
  \footnotesize Orchestration \& Coordination Layer
};

\node[service, minimum width=2.5cm] (output) at (2,-6.5) {\textbf{Output}\\Results};
\node[monitor] (eval) at (8.5,-6.5) {\textbf{Evaluators}\\Monitoring \& Validation};

\draw[->, thick, color=blue!60] (own.south) -- (runtime.north -| own) node[midway, left, font=\tiny] {API};
\draw[->, thick, color=blue!60] (crm.south) -- (runtime.north -| crm) node[midway, right, font=\tiny] {API};
\draw[->, thick, color=blue!60] (chat.south) -- (runtime.north -| chat) node[midway, left, font=\tiny] {API};
\draw[->, thick, color=blue!60] (plane.south) -- (runtime.north -| plane) node[midway, right, font=\tiny] {API};

\draw[->, thick, color=gray!60] (runtime.south -| output) -- (output.north) node[midway, left, font=\tiny] {Results};
\draw[->, thick, color=gray!60] (runtime.south -| eval) -- (eval.north) node[midway, right, font=\tiny] {Metrics};
\draw[->, thick, color=gray!60] (output.east) -- (eval.west) node[midway, above, font=\tiny] {Validate};

\draw[->, dashed, very thick, color=orange!70] (sotopia.south) -- (chat.north) node[midway, right, font=\scriptsize\itshape] {Agent Comm};

\node[draw=gray!70, dashed, very thick, rounded corners=8pt,
      fit=(own)(crm)(chat)(plane)(runtime)(sotopia)(eval)(output),
      inner sep=12mm,
      label={[font=\large\bfseries, fill=white]above:Docker Compose Network}] (docker) {};

\node[anchor=north west] at (docker.south west) {
  \begin{tabular}{l}
    \footnotesize\textcolor{gray}{Host: AWS EC2 t3.large | Ubuntu Linux | Docker 24.0 | Python 3.12}\\
    \footnotesize\textcolor{blue!60}{— Service API} \quad 
    \footnotesize\textcolor{orange!70}{- - Agent Comm} \quad
    \footnotesize\textcolor{gray!60}{— Data Flow}
  \end{tabular}
};
\end{tikzpicture}
}% end resizebox
\caption{Environment architecture overview. The OpenHands Agent Runtime orchestrates OwnCloud, EspoCRM, Rocket.Chat, and Plane via APIs; Sotopia injects agent capabilities into Rocket.Chat; outputs are validated by Evaluators.}
\label{fig:env-arch}
\end{figure}
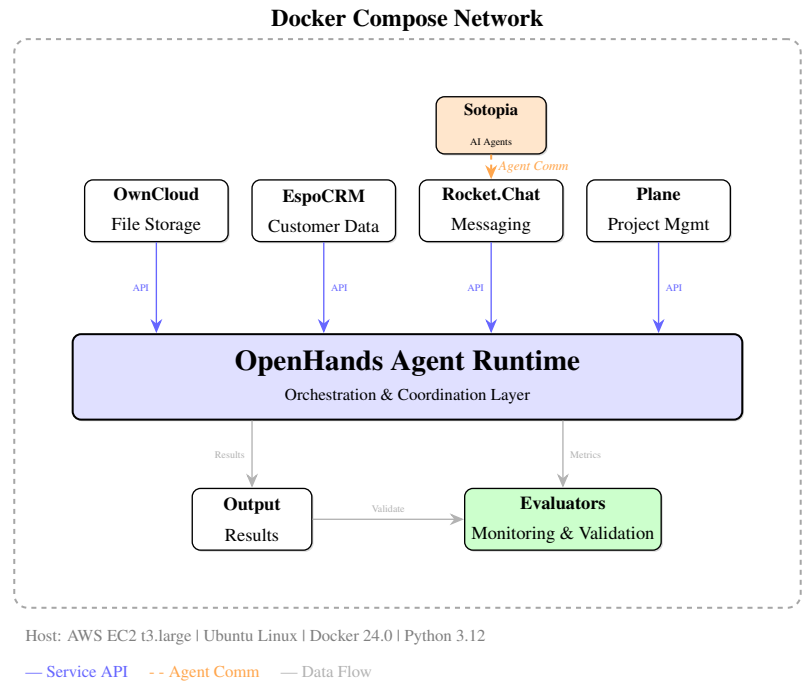
\endgroup

\subsection{Tools and roles}\label{subsec:tools}
At the centre of the environment (Fig.~\ref{fig:env-arch}) is the \emph{OpenHands} agent runtime, which orchestrates a plan–act loop and invokes tools through service APIs. The agent reads client records and attachments from \emph{EspoCRM} (customer data/CRM); retrieves inputs and writes deliverables to \emph{OwnCloud} (file storage via WebDAV); exchanges confirmations and stakeholder updates in \emph{Rocket.Chat} (messaging); and records or checks completion state in \emph{Plane} (project management). \emph{Sotopia} supplies simulated colleagues that interact \emph{only via Rocket.Chat}, providing realistic dialogue. Task artefacts and communications produced by the agent are then consumed by our deterministic \emph{Evaluators} (\texttt{grade\_checkpoints}), which validate file locations and schemas, numerical results, and required messages or tickets by reading from the same services. All components run as Docker containers under a single Compose network with fixed hostnames/ports and named volumes, ensuring reproducible executions across runs and hosts.

\subsection{Relation to TAC architecture}\label{subsec:env-vs-tac}
Other than the additional tasks, data, and improved evaluators, our environment follows TheAgentCompany’s Dockerised design (OpenHands + OwnCloud + Rocket.Chat + Plane) but \emph{adapts the tool mix to finance}: we \emph{add} EspoCRM for client/record management, which improves on TAC by adding realism into benchmarking the financial sector. We \emph{omit} GitLab since software-development tasks are out of scope. Evaluators and the run harness remain TAC-compatible so task logic, paths, and IDs are stable across reruns, while data and tasks are specialised to wealth-management workflows.

\section{The choice of EspoCRM}\label{sec:espocrm-choice}
We selected a Customer Relationship Management (CRM) system to centralise client interactions, data, and workflows so agents operate in a realistic wealth-management setting. In our context, staff must access consistent profiles, contact history, preferences, and transaction logs, to perform their roles reliably. Without a CRM, data fragments across directories and becomes inconsistent. Because TheAgentCompany (TAC) does not include a CRM, we introduced one to provide realism in the assistant’s role and to improve the validity of our evaluation, making results more reflective of real-world performance.

Common CRM options include:
\begin{itemize}
\item Salesforce Financial Services Cloud
\item Microsoft Dynamics 365 Finance and Operations
\item Zoho CRM for Finance
\end{itemize}
These platforms manage customer data efficiently and offer strong advisory tooling; however, they are not open-source and fall outside our project scope. We therefore set out to find a reliable, open-source CRM that is easy to develop with.

We evaluated several open-source solutions, SuiteCRM, OroCRM, and YetiForce, and also considered a custom backend on Supabase. We prioritised lightweight architecture, modularity, RESTful API support, Docker compatibility, and permissive licensing. SuiteCRM proved feature-rich but overly complex and resource-intensive for a proof-of-concept. OroCRM offered enterprise-grade functionality that exceeded our needs and introduced licensing overheads. YetiForce was lightweight but showed instability when containerised and had limited community support. Supabase was attractive as a real-time Postgres backend with built-in authentication and API generation, but it lacked a native CRM interface and would have required significant custom UI development to replicate contact-management workflows.

We chose \textbf{EspoCRM} because it struck the best balance: it provides an out-of-the-box CRM UI, is fully extensible via modules, includes a mature RESTful API, runs smoothly in Docker, and is backed by an active open-source community. We integrated EspoCRM to store client records (including trust/beneficiary tables and attachments) that agents query and reconcile against OwnCloud documents. This addition, layered onto TAC’s orchestration, increases domain realism while keeping runs reproducible and evaluator-friendly.

\section{EspoCRM Deployment: Platform, Configuration, and Integration}\label{sec:espocrm-deployment}
We deployed \textbf{EspoCRM} as part of the TAC Docker stack to make agent evaluation reproducible and realistic. We added an EspoCRM service (with its database) to \texttt{docker-compose.yml}, exposed a fixed host port, and assigned a stable container hostname. These choices give the OpenHands agent and the evaluators a consistent base URL and credentials, so every run queries the same API endpoints and record IDs without per-run reconfiguration.

We configured persistent Docker volumes for the app and database to keep client records, attachments, and IDs stable across runs. This persistence is essential for grading: evaluators can rely on unchanging primary keys and paths when verifying that the agent fetched the correct CRM tables and reconciled them against OwnCloud documents.

We integrated EspoCRM with the rest of the environment through the same host-network access used by TAC (e.g., OwnCloud, Rocket.Chat, Plane). The agent authenticates via EspoCRM’s built-in API auth, and evaluators read only final service state (not transient logs), which keeps checks deterministic.

To ensure smooth agent access, we normalised data encodings to UTF-8 and verified that CRM exports and attachments open cleanly in both the browser and \texttt{pandas}. When needed, we restore the CRM to a known state by re-seeding the named volumes, guaranteeing that results remain comparable over time.

In short, we kept the deployment minimal and evaluation-driven: a small compose extension, stable endpoints, persistent volumes, and UTF-8–clean data. These decisions eliminate configuration drift and make the presence and correctness of CRM data a dependable basis for automated grading.

\section{Task Suite Design}\label{sec:task-suite-design}
\subsection{Preparatory work: Interviews and Research}\label{sec:proprietary-work}
The goal of the preparatory work was to identify realistic, automatable activities performed within wealth-management teams that could be expressed as well-specified tasks with objectively scorable outputs. The assistant layer of the workflow was deliberately targeted, artefacts such as summaries, checklists, draft communications, data pulls, and simple analytics, rather than discretionary investment decisions. This choice increases evaluability (clear acceptance criteria) and reduces ethical/competence risks associated with unqualified financial advice.

We conducted a semi-structured interview with a senior product manager at Aveni Ltd (a firm specialising in Artificial Intelligence for financial services). The interview was scheduled for approximately 1 hour on 29/05/25, for which notes were taken. The structure of the interview was very basic and consisted of (i) What particular jobs within the financial industry has potential to be automated? (ii) What are some of the typical tasks performed in this job? (iii) What are some example tools that are used in this specific field? This interview provided guidance on the exact role to focus on, specific tasks that are expected in this role (e.g. generating portfolio summaries, scheduling meetings, organising documents, and corresponding with clients) and tools that are commonly used in this job, such as document storage, messaging apps, task tracking, and the largest suggestion – a Customer Relationship Management (CRM) tool.

To complement the insights gained from the interview, supplementary research was conducted into the typical responsibilities and workflows of wealth managers’ assistants. This involved reviewing recent literature and industry reports on automation in financial advisory roles, focusing specifically on tasks commonly delegated to assistants, such as preparing client summaries, compliance checking, administrative record-keeping, scheduling, and drafting standardised client communications. From this review, it became clear that assistant-level tasks frequently revolve around structured data management, repetitive administrative processes, and routine client interactions, all activities particularly suited to automation due to their clear inputs, predictable processing steps, and objectively verifiable outputs. These insights aligned closely with the findings from our expert interview and helped confirm the appropriateness of the chosen assistant-level tasks for automated benchmarking within this project.
This investigation was focused on a wealth management assistant rather than a wealth manager for multiple reasons. First, assistant-level outputs are amenable to automated scoring, as they have verifiable ground truth answers which enable repeatable experiments without a panel of domain experts for every run. Second, the academic literature is thin on rigorous agent benchmarks tailored to wealth-management back-office/assistant functions; focusing here adds incremental value rather than repeating trading-strategy or generic Q\&A evaluations [ref]. By deferring discretionary investment judgements to future work, the present study isolates capabilities (information retrieval, synthesis, compliance checklists, communication drafting) that are both useful and objectively checkable.

\subsection{Task Creation Method and Task Descriptions}\label{sec:task-suite-design}

Our suite of 12 task pairs were carefully designed to reflect authentic activities encountered by wealth management assistants, based directly on the guidance and research outlined in the preceding section. Twelve tasks were created, mirroring TAC’s existing financial suite. Each task was structured to clearly define an intent, explicit inputs (such as file paths and CRM endpoints), and objectively verifiable outputs. Additionally, we assigned a difficulty level based on the complexity of the operations involved:

\textbf{Level 1 (Basic Retrieval and Summarisation):}
\begin{itemize}
    \item Tasks at this level involve straightforward retrieval of client-related data and summarisation into a specified format. Examples include extracting transaction histories or asset listings from CRM or document storage, then aggregating or formatting them according to predefined templates. Such tasks typically required minimal processing beyond accurate data collection and organisation.

\end{itemize}

\textbf{Level 2 (Analytical and Computational Processing):}
\begin{itemize}
    \item Tasks classified at Level 2 require analytical processing, such as performing numerical calculations or data reconciliations. Examples include reconciling beneficiary percentages between CRM and trust deeds, or calculating gains/losses using First in First out (FIFO) on trade history, meaning each sale is matched to the earliest remaining purchase lots first, so the gain/loss uses the oldest acquisition prices until the sold quantity is fully allocated. These tasks necessitate both data accuracy and correct application of domain-specific computational rules.

\end{itemize}

\textbf{Level 3 (Synthesis, Communication, and Decision Support):}
\begin{itemize}
    \item These tasks involve the synthesis of information from multiple sources, preparation of structured client communications, and provision of actionable decision support. Representative tasks include analysing quarterly business metrics to formulate concise insights for stakeholder communication via messaging applications, evaluating a colleague’s financial position to determine if savings adequately cover future expenses, and coordinating meetings between clients and financial advisors by aligning multiple calendars. Such tasks require the agent to integrate and interpret diverse data inputs, apply reasoning to reach clearly defined decisions, and produce accurate, coherent, and client-ready outputs, such as scheduled meeting invitations, succinct stakeholder messages, or pass/fail assessments with precise numeric indicators.

\end{itemize}

\begin{center}
\small
\begin{longtable}{@{}p{0.25\textwidth} p{0.38\textwidth} p{0.31\textwidth}@{}}
\caption{Task overview: purpose, sources, and deliverables\label{tab:task_descriptions}}\\
\toprule
\textbf{Task} & \textbf{Purpose} & \textbf{Sources $\rightarrow$ Deliverable} \\
\midrule
\endfirsthead
\toprule
\textbf{Task} & \textbf{Purpose} & \textbf{Sources $\rightarrow$ Deliverable} \\
\midrule
\endhead
\midrule \multicolumn{3}{r}{\itshape continued on next page} \\
\endfoot
\bottomrule
\endlastfoot

\multicolumn{3}{@{}l}{\textbf{Level 1}}\\
\midrule
Net Worth Snapshot
 & Summarise a client’s net worth across assets and liabilities.
 & \emph{OwnCloud} $\rightarrow$ short net-worth note saved. \\
\midrule

\multicolumn{3}{@{}l}{\textbf{Level 2}}\\
\midrule
Asset Aggregation
 & Aggregate cash, investments, and property into a total-assets figure.
 & \emph{OwnCloud} $\rightarrow$ consolidated total-assets note. \\
\midrule
Year-End Tax Summary
 & Compile taxable income components for tax reporting.
 & \emph{OwnCloud} $\rightarrow$ year-end tax summary table. \\
\midrule
Trusts \& Beneficiaries
 & Reconcile trust records with beneficiaries to find discrepancies.
 & \emph{EspoCRM} + \emph{OwnCloud} $\rightarrow$ discrepancy report. \\
\midrule
Tax Threshold Calculation
 & Compute estate-level tax due using bands and rates.
 & \emph{OwnCloud} $\rightarrow$ single-figure tax-due note. \\
\midrule
Pension Projection
 & Project contributions and growth to retirement.
 & \emph{EspoCRM} $\rightarrow$ multi-year projection table (saved to OwnCloud). \\
\midrule
Expense Categorisation
 & Classify spending to reveal saving opportunities.
 & \emph{EspoCRM} $\rightarrow$ categorised transactions export (saved to OwnCloud). \\
\midrule
Capital Gains Computation
 & Calculate gains/losses using FIFO on trade history, .
 & \emph{OwnCloud} $\rightarrow$ capital gains report (saved). \\
\midrule
Portfolio Asset Allocation
 & Summarise portfolio by asset class (percent allocations).
 & \emph{OwnCloud} $\rightarrow$ asset-class breakdown (saved). \\
\midrule

\multicolumn{3}{@{}l}{\textbf{Level 3}}\\
\midrule
Emergency Fund
 & Check if savings cover \(\geq\) six months of expenses.
 & \emph{OwnCloud} $\rightarrow$ pass/fail message posted in \emph{Rocket.Chat}. \\
\midrule
Meeting Scheduling Coordination
 & Align client and advisor calendars and confirm a meeting.
 & \emph{EspoCRM} + \emph{Rocket.Chat} $\rightarrow$ confirmation message; calendar invite saved to \emph{OwnCloud}. \\
\midrule
Quarterly Metrics Analysis \& Communication
 & Analyse quarterly business metrics and inform stakeholders.
 & \emph{OwnCloud} + stakeholder list $\rightarrow$ updates in \emph{Rocket.Chat}; task confirmed in \emph{Plane}. \\

\end{longtable}
\end{center}

Our task suite described in \ref{tab:task_descriptions} collectively addresses the full spectrum of assistant-level competencies, from basic information retrieval to more sophisticated analytical reasoning and client communications. Each task includes clearly defined acceptance criteria, output schema (CSV, JSON, .txt, or structured chat messages), and precise input sources (e.g. EspoCRM endpoints, OwnCloud file locations, or Rocket.Chat user identifiers). Our comprehensive structuring and explicit specification of each task ensures objective and reproducible evaluation, reducing ambiguity and facilitating automated scoring.

\subsection{Checkpoint Design and Description}\label{sec:checkpoint-design}

The purpose of checkpointing is to transform each multi-step task into a sequence of verifiable evaluations that yield precise, interpretable feedback on agent performance. Checkpoints are not exposed to the agent, rather, they are a design aid used to decompose a task before writing the evaluator script and to provide a human-readable account of what the evaluator will grade, this is not strictly necessary but instead we completed this in order to remain consistent with legacy TAC setup. In practice, checkpointing lets us (i) pinpoint the exact sub-step e.g. data retrieval, calculation, or formatting, at which a run breaks down, instead of marking the entire task as failed; (ii) follow TheAgentCompany pattern in which human-readable checkpoints are the basis for an evaluator, and (iii) ensure transparency and reproducibility by documenting unambiguous acceptance criteria in \texttt{checkpoints.md}, making the pipeline auditable and shareable.

For each task we first extract the key features: the input locations and files, the essential processing steps, and the expected deliverables (format and destination), this has to be done manually for every new task we created. We then define a compact set of three to five checkpoints for each to balance diagnostic granularity against unnecessary complexity, this is the main improvement to checkpoints over the prior TAC checkpoints which often has fewer than 3 checkpoints with higher points awarded per checkpoint. This design is inefficient as it obscures where runs fail due to evaluation not covering it (access, computation, or delivery), inflating score variance, and yields limited, non-actionable feedback for improving agents. Our setup then specifies succinct acceptance criteria so that each checkpoint reflects how well the agent performed that particular step.

Across tasks, the acceptance criteria reflect only externally verifiable behaviour and artefacts. Concretely, checkpoints fall into five practical dimensions: \emph{input location \& access} (the agent finds and opens the correct files/records or endpoints); \emph{computation correctness} (e.g. sums, projections, or FIFO gains); \emph{artefact conformance} (the correct file exists at the described path with the correct name and required schema/template); \emph{communication actions} (well-formed Rocket.Chat DMs or calendar invites saved to OwnCloud); and \emph{tool state updates} (e.g. marking a Plane issue as done), this is an improvement on TAC. Importantly, there is no checkpoint for internal “data parsing”; evaluators judge only observable outputs and side-effects from the agent's 'trajectory'. 

\subsection{Checkpoint Design}\label{subsec:checkpoints}

\textit{What is inherited vs.\ provided} The overall approach, checkpointed, post-hoc grading; and the \texttt{Result}/\texttt{Checkpoint} schema follows TheAgentCompany (TAC). Our main contribution in addition to 12 new checkpoints for wealth management is finance-specific checkpoint definitions and acceptance criteria, the finer granularity (3–5 checks per task with small point weights, allowing more effective analysis of what sub-steps of the task the agent may have struggled with).

\textbf{Illustrative example (Level 1: Net Worth Snapshot).}\vspace{0.1em}

As an example, the \emph{Net Worth Snapshot} task awards four points across three checks: (i) the agent must read the client’s \texttt{assets} and \texttt{liabilities} inputs from OwnCloud (1 point possible); (ii) compute the correct totals (1 point possible); and (iii) write a one-line deliverable \texttt{net\_worth\_snapshot.txt} in the client’s folder with the exact format \texttt{Net worth: £<value>} (2 points possible). The evaluator verifies existence and location of the output, parses and validates the numeric value against a recomputation of the inputs, and rejects any extra text or misformatting. This pattern, \emph{inputs read} $\rightarrow$ \emph{numerics right} $\rightarrow$ \emph{deliverable right place/format}, is reused across the suite.\vspace{1em}

A checkpoint passes when its corresponding evaluator returns \texttt{True}. Verifiers fall into four recurring types: (1) \emph{location/schema}: the artefact exists at the expected path and matches the required columns or message template; (2) \emph{numeric correctness}: values recomputed by the grader match the agent’s output within exact equality (for integers) or a tiny tolerance (for floats); (3) \emph{communication}:  correctness is verified with an existing TAC LLM-as-a-judge over the Rocket.Chat DM (yes/no predicate on the transcript), following the prior TAC setup, whereas other checks are deterministic (regex on the action trace and file existence); (4) \emph{state transition}: a Plane issue reaches the expected state. Each task defines a point budget and maps each passed checkpoint to its share; the overall task score is the sum of passed points.\vspace{1em}

Checkpoints are evaluated \emph{post-hoc}, once the agent finishes. The grader inspects service state via APIs/WebDAV and, where helpful, consults the execution trajectory for evidence (e.g., URLs followed). There is no streaming or per-step polling; instead, checkpoints are designed so that final service state is sufficient to determine success. By default, checkpoints are \emph{order-agnostic}: any check can pass regardless of when in the run its evidence was produced, though a natural order usually surfaces. For tasks that logically hinge on a final deliverable, a light scoring strategy awards “make-good” credit to upstream checkpoints when the final one is complete (e.g., if the final CSV is correct, the “read inputs” checkpoint is credited even if a trajectory keyword is missing). This avoids penalising superficial ordering artefacts while still rewarding the intended outputs.\vspace{1em}

Checkpoints target \emph{deliverable and process correctness}, not stylistic choices or intermediate scratch work; a task can succeed even if the agent explored dead ends, provided the final artefacts and required communications are correct.

\subsection{The Creation of Automatic Evaluators}\label{sec:evaluators}

\textbf{What is an evaluator?} In this work, an \emph{evaluator} is the deterministic Python script that implements the acceptance criteria for a task’s checkpoints. It inspects final service state (e.g., files on OwnCloud via WebDAV, Rocket.Chat transcripts via API, task trajectory, Plane issue state) and returns boolean pass/fail for each checkpoint. By contrast, a \emph{metric} aggregates those boolean outcomes (and optionally cost/efficiency signals) into reported scores: per-task points (sum of passed checkpoints), suite-level accuracy (share of available points achieved), and secondary metrics such as API cost or step-count. Put simply: evaluators \emph{decide} if a specific requirement was met; metrics \emph{summarise} those decisions for comparison.

\textbf{Why heterogeneous outputs are hard.} Each task yields a different output type, and naïve exact matching is brittle. Numerical deliverables are relatively straightforward: the evaluator recomputes the quantity from the declared inputs in Python and compares using exact equality for integers or small tolerances for currency/percentages. Text deliverables require exact phrasing and location, and this is strictly specified in the wealth-management \texttt{task.md} files (e.g., one, line files with mandated templates), which we verify with strict regex plus path checks. CSV deliverables are the most challenging: byte-for-byte equality often fails under benign differences (row order, header case, separators). To avoid false negatives, our CSV checks (i) assert required column presence and, where the spec demands it, exact column \emph{order}; (ii) normalise numeric \emph{inputs} (e.g., strip currency symbols, commas, whitespace) before recomputation; and (iii) compare value columns against a recomputed ground truth, treating row order as irrelevant unless ordering is explicitly part of the specification. Human-facing communications (e.g., Rocket.Chat DMs) are the only place an LLM-as-a-judge is used (this can also use strict regular expression as a backup), and then only to answer a yes/no predicate over the transcript; all file-path, schema, and numeric checks remain deterministic.

\textbf{Illustrative example (Year-End Tax Summary).} This evaluator has four deterministic checks (one point each). First, it looks for positive evidence in the trajectory that the agent accessed all three source files (\texttt{income.csv}, \texttt{gains.csv}, \texttt{dividends.csv}) under the correct client directory. Second, it recomputes expected totals by reading those sources, stripping currency symbols/commas, summing the relevant columns, and comparing to the report’s \texttt{Summary} rows within a \(10^{-6}\) tolerance; row order is irrelevant because the check filters \texttt{Section=Summary} and indexes by \texttt{Item}. Third, it verifies the report exists at the specified path. Fourth, it enforces exact schema and formatting, columns \([\texttt{Section}, \texttt{Item}, \texttt{Amount (£)}]\), no placeholder \texttt{x}, and all amounts matching a strict two-decimal regex. No LLM is used in this evaluator; it is fully programmatic. The scoring strategy uses \texttt{bonus\_for\_completing\_final} so a fully correct final artefact can credit upstream access where appropriate.

\textbf{What follows TAC vs.\ what we changed.} The general pattern, checkpointed, post-hoc grading and the \texttt{Result}/\texttt{Checkpoint} schema, follows TheAgentCompany (TAC). However, several TAC pitfalls motivated changes for finance tasks. First, some TAC evaluators awarded credit for weak signals (e.g., a file’s mere existence or a keyword in the trajectory); we require positive evidence of \emph{correct} artefacts and recomputed correctness. Second, coarse, high-weight checkpoints made scores less diagnostic; we use 3–5 smaller checkpoints per task so failures localise to access, computation, formatting, or delivery. Third, exact-string comparisons caused spurious failures; we added normalisation and tolerance for numbers and CSVs while keeping outputs strict where the spec demands exactness (file names, schemas, phrasing, and two-decimal amount formatting). Fourth, TAC sometimes relied on trajectory text over service state, this can skew evaluation by providing points for the mere mention of a location; our evaluators prefer API/WebDAV state (as well as disabling the default Rocket.Chat reset to preserve transcripts), consulting the trajectory only when it adds corroboration. Finally, LLM-judging is constrained to message-content verification; all other checks are programmatic. These changes improve validity (fewer false positives from weak evidence), reliability (fewer false negatives from harmless formatting), and diagnostic value (clearer attribution of failure mode).

\textbf{How evaluators are created.} For each task, we first write a short, human-readable checkpoint spec, then implement one \texttt{@grader} function per checkpoint. Each function queries the relevant service or file system, performs any needed normalisation, and returns a boolean. The \texttt{eval.py} harness runs these functions \emph{after} the agent completes and composes them into a \texttt{Result}. Most tasks use simple sum-of-parts scoring; when a final deliverable certifies upstream work (e.g. a fully correct CSV implies inputs were read), a light “make-good” strategy such as \texttt{bonus\_for\_completing\_final} credits the first checkpoint to avoid penalising missing trajectory breadcrumbs.

\textbf{Why this is necessary.} Finance tasks mix arithmetic precision with document discipline and light workflow actions. Evaluators that are too lenient (existence checks) inflate scores without real capability; evaluators that are too strict (byte-exact CSV equality) punish harmless differences. The tightened yet tolerant design here aims for the middle ground: deterministic where possible, minimally LLM-assisted only for chat semantics, and explicitly tuned to the output type so that scores are both fair and reproducible. This is more effective in evaluation by allowing evaluators to correctly grade how correct the agents actions are, providing valuable understanding on common pitfalls agents experience.

\section{Process of Modifying Task Autonomy}\label{sec:task-specificity}

We reworked each task along a spectrum from \emph{high autonomy} (brief, outcome-only) to \emph{low autonomy} (schema-aware, tool- and I/O-constrained). The aim was not just to change the wording, but to test how instruction granularity affects agent \emph{accuracy}, \emph{efficiency}, and \emph{evaluability}. High-autonomy briefs encourage exploration but often trigger tool wandering, re-downloading, or partial deliverables; low-autonomy briefs constrain sources, methods, and outputs so evaluators can verify results deterministically.

Concretely, we applied a three-pass refinement: intent capture (state the real-world outcome), constraint surfacing (which tools are in scope, where the data must come from, and what \emph{not} to do), and I/O formalisation (required columns/phrasing, output naming, and observable delivery actions). Each pass removes ambiguity that previously led to drift and non-reproducible outputs. 

\textbf{High vs.\ low autonomy (what changed and why).}
\begin{itemize}
  \item \emph{Data provenance.} \textbf{High:} “Find the relevant data.” \textbf{Low:} Name the canonical sources (specific CRM tables and the corresponding deed in the document store), prefer local copies if present, and \emph{forbid} fabricating data or redundant re-downloads. \textbf{Why:} prevents hallucinated inputs and locks evaluation to known ground truth.
  \item \emph{Tool binding.} \textbf{High:} tool choice implicit. \textbf{Low:} declare the systems the agent must use (OwnCloud, EspoCRM, Rocket.Chat, Plane) and the allowed access route (local mount/API). \textbf{Why:} reduces detours and yields observable evidence.
  \item \emph{Output canon.} \textbf{High:} “Produce a report.” \textbf{Low:} fix filename pattern, required columns/phrasing, and where it must be written or sent. \textbf{Why:} enables strict, programmatic checks (schema, wording, location).
  \item \emph{Computation rules.} \textbf{High:} unspecified. \textbf{Low:} state methods (e.g., FIFO for gains, banded tax, compounding convention). \textbf{Why:} removes ambiguity that otherwise causes “nearly correct” but unverifiable answers.
  \item \emph{Delivery.} \textbf{High:} implicit. \textbf{Low:} require a verifiable action (upload to a known path, mark a Plane issue, or send a formatted DM). \textbf{Why:} closes the loop so evaluators can confirm completion without human review.
\end{itemize}

\begin{center}
\renewcommand{\arraystretch}{1.25}
\footnotesize
\begin{longtable}{|p{3.2cm}|p{5.6cm}|p{5.6cm}|}
\caption{Accurate comparison for the “Beneficiary discrepancies” task: the high-autonomy brief names tools and files but leaves method and policies open; the low-autonomy brief constrains access (local-first), enumerates steps, tightens schema (adds \texttt{Difference}), and mandates upload.}
\label{fig:autonomy-contrast-corrected}\\
\hline
\textbf{Aspect} & \textbf{High autonomy brief} & \textbf{Low autonomy brief} \\
\hline
\endfirsthead
\hline
\textbf{Aspect} & \textbf{High autonomy brief} & \textbf{Low autonomy brief} \\
\hline
\endhead

Goal wording & Narrative: “find discrepancies in trust beneficiaries … produce a CSV.” 
& Imperative: “reconcile beneficiary percentages” with an explicit step list. \\
\hline
Sources named & Yes: EspoCRM URL and document names (\texttt{beneficiaries.csv}, \texttt{trusts.csv}); deed path in OwnCloud. 
& Yes, plus \emph{local-first} paths for CRM tables and the deed. \\
\hline
Access policy & Not specified (agent can download as it sees fit). 
& \textbf{Local-first}; “fetch once” if missing; \textbf{do not re-download duplicates}. \\
\hline
Process steps & Implicit: “connect, download, query, reconcile, export.” 
& \textbf{Explicit 1–5}: extract \texttt{trust\_id}; record CRM \%s; read deed \%s; name-match; write per-trust CSV. \\
\hline
Output schema & Columns: \texttt{Beneficiary\_Name}, \texttt{CRM\_Percent}, \texttt{Deed\_Percent}. 
& \textbf{Adds} \texttt{Difference} column; states policy: include all beneficiaries or only discrepancies (choose one consistently). \\
\hline
Output path & Save to client folder with pattern \texttt{<trust\_id>\_beneficiary\_discrepancies.csv}. 
& Same pattern and folder \textbf{plus} explicit \emph{upload} to OwnCloud after creation. \\
\hline
\texttt{trust\_id} handling & Mentions \texttt{trust\_id} comes from CRM but not how to derive in steps. 
& \textbf{Step 1} is “extract \texttt{trust\_id} from \texttt{trusts.csv}.” \\
\hline
Anti-drift constraints & None stated (no local-first or duplication rule). 
& \textbf{Explicit} no re-downloads; read local copies; consistent inclusion policy. \\
\hline
\end{longtable}
\end{center}

\noindent \textbf{Why we did this.} Tightening autonomy is not about making tasks easy; it is about isolating whether agents can \emph{reliably execute} realistic workflows when the business rules (sources, schema, delivery) are unambiguous. This improves validity (grounded inputs), reliability (reproducible outputs), and diagnostic power (failures localise to access, computation, formatting, or delivery), letting us cleanly measure the effect of autonomy on performance.

\section{Data Generation}\label{sec:data-generation}

\paragraph*{Authorship and scope.}
We created the finance-domain corpus needed to run and grade the benchmark. In total, we produced \emph{twenty-eight} artefacts across the stack: most are CSV and TXT files seeded into OwnCloud, with a smaller set placed in EspoCRM (including three calendar PDFs). These datasets span five named client profiles plus one colleague used for scheduling. We built on TAC’s legacy seed (NPC accounts, baseline channels, volumes), so we did \emph{not} recreate those; instead, we added only the finance-specific data required for our tasks.

\paragraph*{How the data was generated.}
For each task family, we drafted a brief in natural language (heading, general  formatting, units, simple rules) and used \emph{ChatGPT o4-mini-high} to synthesise candidate tables/documents from that description. During pilots we observed failures to open files caused by mixed encodings and line endings (e.g., UTF-8 with BOM, Windows-1252, CRLF). We therefore re-encoded \emph{all} artefacts to UTF-8 \emph{without} BOM, normalised newlines to LF, and standardised a single CSV dialect (comma delimiter, RFC-style quoting). This eliminated spurious I/O errors in the browser, Python/pandas, and WebDAV paths, making parsing behaviour reproducible across services and reruns.

\paragraph*{Non-relevant scaffolding.}
To keep the environment as realistic as possible to a real work environment, we added a small amount of \emph{non-relevant} filler data that is not intended for the agent to access (e.g. 2 extra irrelevant calendar files in EspoCRM). This non relevant data we created was kept minimal because TAC already provides a large suite of data that is inherently irrelevant to our new suite. 

\paragraph*{File-path discipline.}
We fixed canonical paths and endpoints so evaluation remains deterministic and realistic. OwnCloud follows a client-centric layout such as:

\texttt{/Finance\_Documents/Clients/\textless Client\textgreater/\,\dots}, while EspoCRM objects are reachable at stable host/port routes. For both task variants we mirror the same artefacts in the agent’s local workspace, but explicitly state “read locally; do not re-download,” for only the low autonomy variant. High-autonomy variants keeps this vague, but the paths remain stable so evaluators can locate ground truth and verify deliverables without brittle search heuristics.

\paragraph*{Why this differs from TAC’s samples.}
TAC provides an excellent harness but its example data is not finance-focused and rarely couples tools. Our corpus intentionally creates \emph{cross-service dependencies} (e.g., reconciling CRM tables against a deed in the document store) within a \emph{closed world} where every task is solvable from the seeded services or their local mirror. This improves validity (agents must use the intended tools), reliability (stable paths/IDs across runs), and diagnostic power (failures localise to access, arithmetic, formatting, or delivery), while remaining evaluator-friendly by avoiding unnecessary byte-identical requirements.

\paragraph*{Contribution.}
In summary, we authored the finance datasets, designed the directory structure and IDs, re-encoded all artefacts to UTF-8/LF for robust agent access, and seeded the content into OwnCloud and EspoCRM at fixed paths. TAC’s orchestration is retained; the finance data, its cross-service links, encoding normalisation, and pathing conventions are ours. This combination makes the benchmark realistic for wealth-management workflows and reproducible for grading.

\section{NPCs and Services Creation}\label{sec:npc-and-services}

A reproducible agent environment needs more than static files; it must also provide stable chat partners and line-of-business apps. For each task we ship three services: Rocket.Chat (dialogue), Plane (issue tracking), and OwnCloud/EspoCRM (documents/records). We start them with a project-wide \texttt{docker-compose.yml} and mount named volumes preseeded with state. With one command (\texttt{docker compose up}) we recreate exactly the world assumed by the evaluator.

\textbf{Rocket.Chat and Sotopia NPCs}

We create each NPC as a normal Rocket.Chat account via the web UI so IDs and channel membership stay fixed; the \{\texttt{username},\texttt{password}\} live in \texttt{npc\_credentials.json}. We define persona/role/prompts in Sotopia’s \texttt{npc\_definition.json} and load them into Sotopia Redis with an idempotent helper. Task-local context comes from \texttt{scenarios.json}. We copy these JSONs into the image in the \texttt{Dockerfile}, and \texttt{init.sh} starts a Sotopia worker that logs in with the stored credentials and streams OpenAI completions as live replies. We avoid resets at startup, so history, unread flags, and presence behave like production.

\textbf{Plane workspace}

We keep a single open issue, “Analyze Q2 Metrics,” on a named PostgreSQL volume referenced in \texttt{docker-compose.yml}. \texttt{plane\_seed.tgz} is a DR fallback only; we do not unpack it during normal runs.

\textbf{OwnCloud and EspoCRM}

We place all CSVs/PDFs/templates under \texttt{/Finance\_Documents}, matching task paths. We uploaded the content once via the OwnCloud web client, set ACLs, and treat the \texttt{ocdata} volume as authoritative. We populated EspoCRM via the admin UI so primary keys match the evaluators; its database and uploads volumes are the sources of truth. \texttt{owncloud\_seed.tgz} and \texttt{espocrm\_seed.tgz} are cold-storage snapshots for recovery.

\textbf{Network topology}

All containers join the \texttt{theagentcompany\_default} bridge network and expose DNS aliases (\texttt{crm}, \texttt{owncloud}, \texttt{chat}, \texttt{plane}). We publish host ports 8080 (EspoCRM), 8092 (OwnCloud), 3000 (Rocket.Chat), and 8091 (Plane); we run the OpenHands runtime with \texttt{--network host} so it reaches each service directly.

\textbf{Resulting invariants}

Because live volumes persist across container restarts, agents can rely on four guarantees:
(1) NPC accounts always exist and answer direct messages;
(2) the Plane ticket is present exactly once and starts in the Open state;
(3) every file referenced in a task is already in place; and
(4) database primary keys and slugs used by evaluators never drift.

If any store is damaged, we stop the stack, extract the relevant \texttt{.tgz} into the named volume, and relaunch \texttt{docker compose}. This restores the environment bit-for-bit in seconds without redundant unpacking on normal runs, yielding a closed-world testbed where every dependency (persona, ticket, document) is exactly where the agent and evaluator expect it, so scores reflect reasoning, not configuration drift.

\section{Performing Evaluation with OpenHands}\label{sec:evaluation}

We run all experiments with the stock OpenHands runner from The Agent Company (TAC) and tighten build/initialisation/scoring so a fresh clone plus two environment variables (OpenAI API key, server hostname) reproduces every table. Python~3.12 with Poetry installs the few extra dependencies. Graders are deterministic; an LLM grader is used only for genuinely free-form text.

For each of the 24 tasks (high/low-autonomy), we provide a short \texttt{Dockerfile} layered on TAC’s task base image, pin \texttt{pandas}/\texttt{numpy}, copies certain seed artefacts locally, and include an AES-encrypted \texttt{evaluator.py}. The evaluator defines 3–5 \emph{checkpoints} (“steps”) as programmatic predicates over files, numbers, and side-effects (e.g. Plane status, Rocket.Chat DMs). Tasks that touch Rocket.Chat and/or Plane add a small \texttt{init.sh} to resolve hostnames, wait for all services to become healthy, load NPCs, and deliberately \emph{skip} TAC’s default resets so chat and Plane state persist. A minimal \texttt{dependencies.yml} enumerates external services so that run\_eval.py can pre-warm health probes; developer niceties (e.g., a tiny \texttt{Makefile}) stay out of the automated path.

We launch via \texttt{run\_eval.py} (OpenHands): the wrapper takes the built task image, reads creds from \texttt{config.toml}, picks an output dir, and uses host networking so the agent reaches EspoCRM (:8080), OwnCloud (:8092), Rocket.Chat (:3000), and Plane (:8091). The agent receives the task prompt, runs until a \emph{finish} action, and the action–observation trace is written to \texttt{traj\_.json}. The evaluator is then decrypted in-container to emit \texttt{eval\_.json} with: (i) a binary \emph{full completion} (all checkpoints pass) and a weighted \emph{partial} score; (ii) string checks (normalised exact match and regex for required phrases); (iii) numeric tolerances (max\,$10^{-6}$ absolute or 1\% relative); (iv) CSV grading by schema/keys (order-insensitive) with values recomputed and compared within tolerance, never byte-for-byte;  and (v) REST checks confirming side-effects.

\FloatBarrier
\chapter{Results \& Evaluation}\label{chap:results-&-evaluation}

\section{Experiment 1 – New Finance Tasks vs.\ Original TAC}\label{sec:experiment1}

\noindent
In these experiments we compare two \emph{distinct} task suites: the 12 original TAC tasks and the 12 new wealth-management tasks (high-autonomy prompts), and another comparison of high/low autonomy prompts. We run the OpenHands agent configured with \textbf{GPT-4o}. For each task we record three outcomes: (i) the \emph{percent of checkpoints passed} (how far the agent progressed on the task), including how many were fully completed, (ii) \emph{API cost} (USD), and (iii) \emph{mean step-count} (number of actions), step-counts are correlated but not directly proportional to cost. The plots below show distributions across tasks within each suite.

\begin{figure}[H]
  \centering
  \makebox[\textwidth][c]{%
    \includegraphics[height=0.27\textheight,keepaspectratio]{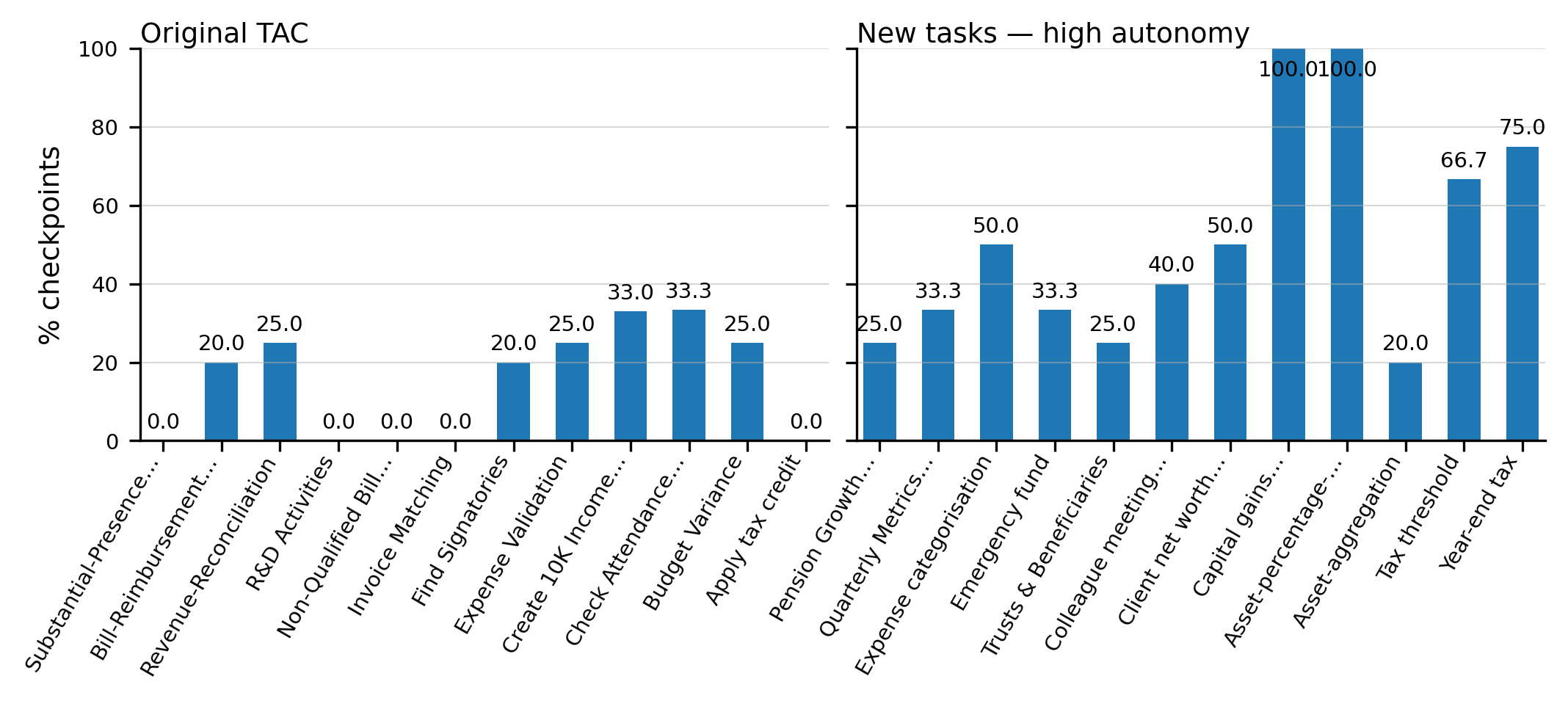}%
  }
  \caption{Experiment 1: \% Checkpoints passed. Left: Original TAC tasks (12). Right: New tasks (high autonomy, 12).}
  \label{fig:exp1-checkpoints}
\end{figure}

\begin{figure}[H]
  \centering
  \makebox[\textwidth][c]{%
    \includegraphics[height=0.27\textheight,keepaspectratio]{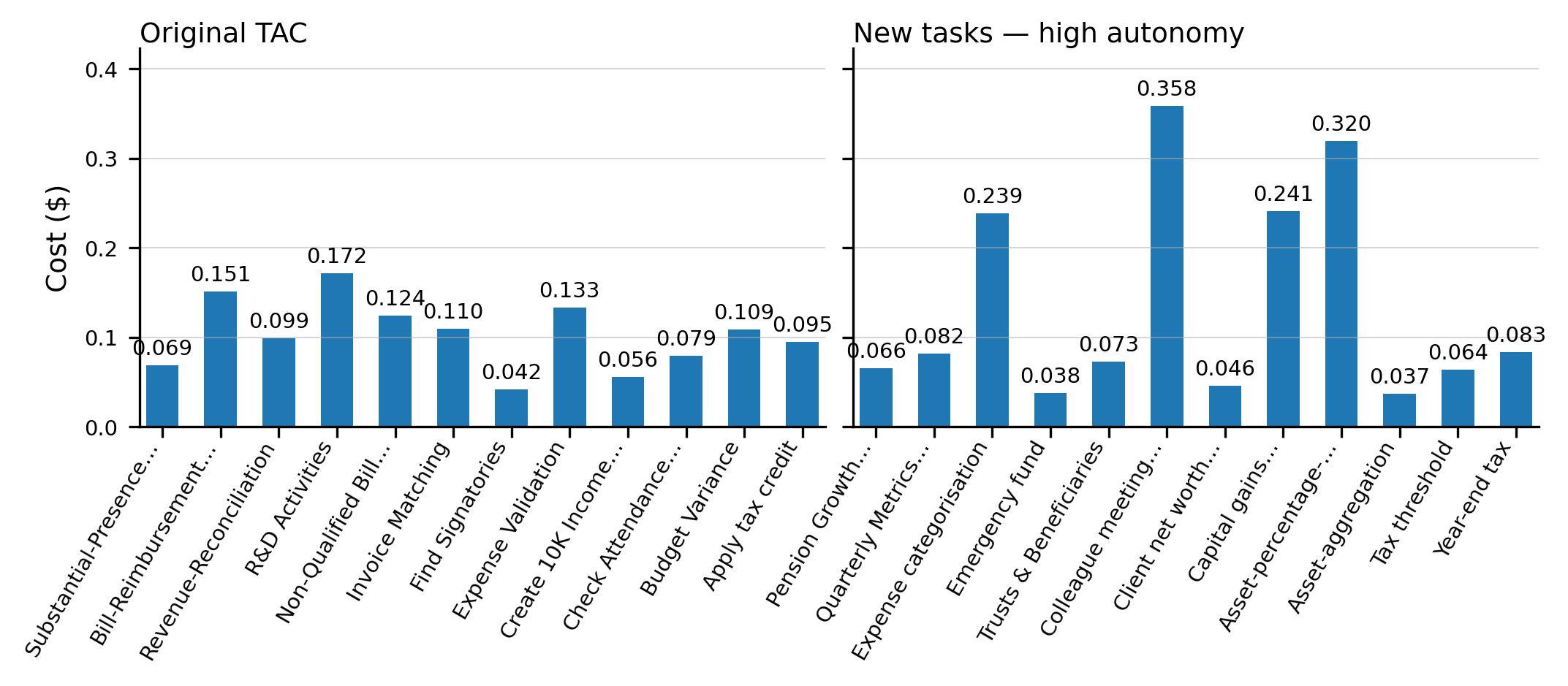}%
  }
  \caption{Experiment 1: Cost distribution. Left: Original TAC tasks (12). Right: New tasks (high autonomy, 12).}
  \label{fig:exp1-cost}
\end{figure}

\begin{figure}[H]
  \centering
  \makebox[\textwidth][c]{%
    \includegraphics[height=0.27\textheight,keepaspectratio]{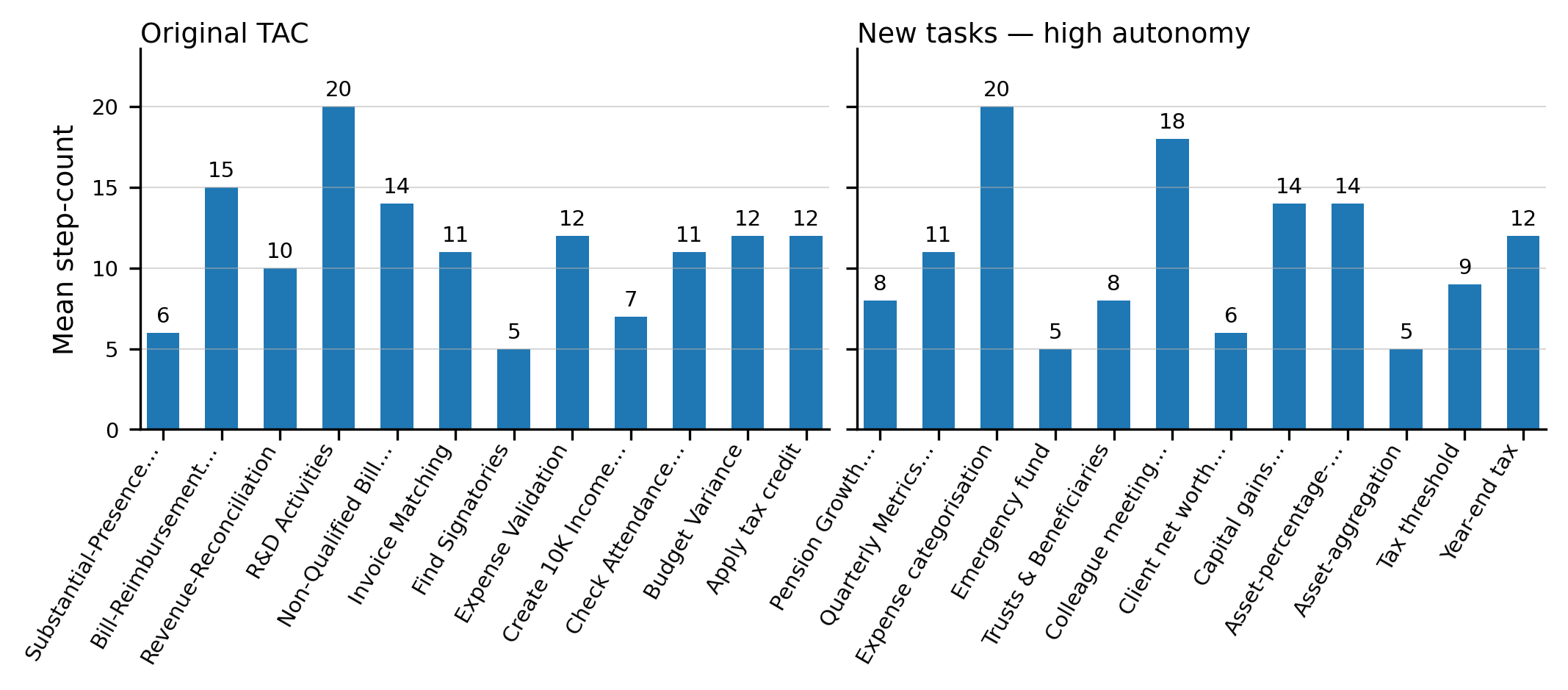}%
  }
  \caption{Experiment 1: Step-counts. Left: Original TAC tasks (12). Right: New tasks (high autonomy, 12).}
  \label{fig:exp1-stepcount}
\end{figure}

\noindent
Figures~\ref{fig:exp1-checkpoints}–\ref{fig:exp1-stepcount} provide a visual overview of Experiment~1.  
Each figure shows \emph{distributions} for two distinct suites: the 12 original TAC tasks versus the 12 new wealth-management tasks with high-autonomy prompts.

\noindent Accuracy rises sharply on the new suite: most TAC tasks lie between 0–33\,\%, whereas the new set includes several tasks above 50\,\% and two at 100\,\%, consistent with Table~\ref{tab:task-comparison} and the distribution in Fig.~\ref{fig:exp1-checkpoints}.

\noindent Costs are broadly comparable (Fig.~\ref{fig:exp1-cost}). A few new-suite outliers (e.g., \emph{Colleague meeting setup}, \emph{Asset-percentage account}) nudge the mean upward.

\noindent Step-counts are similar (Fig.~\ref{fig:exp1-stepcount}); most runs fall between 8 and 14 actions, so higher accuracy on the new suite is not explained by longer trajectories.\vspace{1em}

\noindent\textit{Table~\ref{tab:task-comparison}} aggregates the headline metrics. “Resolved” counts fully completed tasks. “\% Checkpoints Passed” is total checkpoints passed divided by total checkpoints available. “Mean Step-Count” is the average number of actions per run. “Cost (\$)” is the mean API cost per run.

\begin{table}[H]
\centering
\caption{Averages: original TAC finance tasks vs.\ new wealth-management tasks (high-autonomy prompts)}
\label{tab:task-comparison}
\begin{tabular}{|l|c|c|c|c|}
\hline
\textbf{Suite} & \textbf{Resolved} & \textbf{\% Checkpoints Passed} & \textbf{Mean Step-Count} & \textbf{Cost (\$)} \\
\hline
Original TAC (12 tasks) & 0/12 & 8/54 = 14.8\% & 11.3 & 0.1033 \\
New WM (HA, 12 tasks)   & 2/12 & 24/49 = 49.0\% & 10.8 & 0.1373 \\
\hline
\end{tabular}
\end{table}

\subsection{Quantitative Analysis}\label{sec:ex1quantitative}

Table~\ref{tab:task-comparison} and Figures~\ref{fig:exp1-checkpoints}–\ref{fig:exp1-stepcount} show that the new wealth-management suite raises \emph{measured} accuracy without making runs longer. Across the twelve legacy TAC tasks, the agent resolved none and passed 8/54 checkpoints (14.8\%). On the new (high-autonomy) suite, it resolved 2/12 tasks and passed 24/49 checkpoints (49\%). Mean step-count is essentially unchanged (11.3~$\rightarrow$~10.8), and mean API cost rises by \$0.034 (about 33\%), largely due to spreadsheet parsing in richer finance tasks. These differences reflect \emph{improved checkpoints and evaluation} - awarding credit for more intermediary steps (Section~3.4.4), rather than simply “easier” tasks: more of the agent’s partial progress now converts into objectively verifiable checkpoint credit.

\paragraph*{Difficulty tiers:} The new suite was designed with three tiers \ref{sec:task-suite-design}. Aggregating by tier on the high-autonomy runs:
\begin{itemize}
  \item \textbf{Level~1 (retrieval/summarisation):} 2/4 checkpoints (50\%) on \emph{Client net-worth snapshot}; 0/1 tasks resolved; mean 6 steps; mean cost \$0.046.
  \item \textbf{Level~2 (analysis/computation):} 17/31 checkpoints (54.8\%); 2/8 tasks resolved (\emph{Capital gains}, \emph{Asset-percentage}); mean 11.3 steps; mean cost \$0.140.
  \item \textbf{Level~3 (synthesis/communication):} 5/14 checkpoints (35.7\%); 0/3 tasks resolved; mean 11.3 steps; mean cost \$0.160.
\end{itemize}
Two patterns matter. First, higher tiers are \emph{harder}: Level~3 passes far fewer checkpoints than Level~2, despite virtually identical step-counts. Second, costs climb with tier (driven by chat/upload loops in \emph{Colleague meeting setup}), indicating that communication/delivery, not arithmetic, is the dominant friction.

\paragraph*{Task-level contrasts.} Some outcomes are expected (\emph{Capital gains} and \emph{Asset-percentage} both hit 100\% under high autonomy once the files are reachable), while others surface remaining weaknesses. \emph{Asset aggregation} passed only 1/5 checkpoints (20\%) because the agent never opened the three source CSVs, an access/coordination failure, not a calculation one. \emph{Client net-worth snapshot} reached 50\% but did not resolve, showing that even Level~1 retrieval tasks still fail on delivery/formatting constraints when grading is strict. These examples illustrate that the new suite remains challenging while crediting genuine, schema-conformant progress.

\paragraph*{What the comparison shows.}
Versus TAC, the new suite: (i) \emph{reduces spurious credit} by requiring API/WebDAV evidence of correct path/schema/placement (e.g., Year-end Tax headers; Trusts \& Beneficiaries filename), eliminating “keyword-only” passes (cf. clearer separation in Fig.~\ref{fig:exp1-checkpoints}); (ii) \emph{spreads difficulty}, Level~2 improves while Level~3 remains low (\S\ref{sec:ex1quantitative}); (iii) \emph{raises measured accuracy at similar search depth}, checkpoint gains in Table~\ref{tab:task-comparison} occur with near-unchanged step-counts (Fig.~\ref{fig:exp1-stepcount}); and (iv) \emph{converts real computation into credit} via allowed local mirrors (e.g., \emph{Capital gains}, \emph{Asset-percentage}) instead of zeroing runs for transient auth failures.
This supports the research question: it \emph{is} possible to build an effective, reproducible task battery that reflects a wealth-manager’s assistant role and yields meaningful measures of an LLM agent’s fitness for that role. The suite is not easier; it is \emph{better instrumented}, meaning the tasks keep hardness (Level 3 remains unsolved) but yield stronger, verifiable signals (grading localised outputs if none are found in services and ground-truth recomputation), so failures localise to access/delivery rather than being masked by brittle grading. A good benchmark must retain headroom, having only 3 of 12 tasks at 100\% keeps the suite discriminative and useful for tracking future improvements, rather than saturated by trivially solved items.

\vspace{1em}

\subsection{Qualitative Analysis}\label{sec:ex1qualitative}

Failure modes are grouped into a small taxonomy and compare how often they surface in the original TAC suite versus the new wealth-management (high-autonomy) suite. References to Table~\ref{tab:task-comparison} and Figures~\ref{fig:exp1-checkpoints}–\ref{fig:exp1-stepcount} provide the quantitative backdrop.

\paragraph*{Error taxonomy (with representative examples).}
\begin{itemize}
  \item \textbf{Access \& authentication failures (dominant).} In TAC, OwnCloud/Rocket.Chat/Plane routinely returned login pages or 404s by agents querying \emph{unauthenticated}, non-existent static URLs instead of  authenticated export/API endpoints, returning errors rather than files, preventing any real inputs (\emph{Apply tax credit}, \emph{R\&D activities}, \emph{Invoice matching}). In the new suite this persists but is partially mitigated where files are seeded locally, allowing computation to still be evaluated even when cloud auth fails (\emph{Capital gains}, \emph{Asset-percentage}).
  \item \textbf{Data provenance problems (mocked or fabricated inputs).} When access failed, TAC runs often created synthetic data, when not able to find real data (\emph{Expense validation}, \emph{Check attendance}). The new suite still shows this on access-blocked tasks (\emph{Pension growth}, \emph{Asset aggregation} under high autonomy), but less so due to real files being reachable more often.
  \item \textbf{Delivery/transfer failures (uploads \& messaging).} Both suites show repeated misuse of raw \texttt{PUT} instead of WebDAV and Rocket.Chat DMs that never send (\emph{Year-end tax}, \emph{Colleague meeting setup}); the new suite makes these delivery errors more visible because upstream computation more often succeeds.
  \item \textbf{Pathing \& schema conformance.} TAC frequently saved outputs to the wrong location or with missing columns (\emph{Budget variance}), collapsing otherwise correct logic at grading time. Interestingly the new suite exhibits this less frequently but does not eliminate it.
  \item \textbf{Analytical logic errors (less common but important).} TAC: ordering bug in travel-day math (\emph{Substantial presence}). New suite: band-capping/double-count mistakes in tax (\emph{Tax-thresholds}). Where data are available, the suite surfaces genuine reasoning faults rather than mere I/O failures.
\end{itemize}

\paragraph*{Cross-suite contrast.}
In TAC, runs typically died at the \emph{access} boundary, forcing fabrication or early aborts; few tasks reached business logic or delivery. In the new suite, seeded files move more attempts into the compute and deliver stages: successes appear when inputs are present (\emph{Capital gains}, \emph{Asset-percentage}), while remaining failures concentrate mostly on \emph{delivery} (uploads/DMs) and occasional domain logic (\emph{Tax-thresholds}). Thus, the new tasks remain challenging but are better instrumented: they credit real progress and localise residual weaknesses.

\paragraph*{Difficulty signal.}
 
The tiered design behaves as intended. Level~1 retrieval tasks still fail on strict placement/format when auth is flaky; Level~2 computational tasks succeed when inputs load and expose arithmetic competence (\emph{Capital gains}); Level~3 synthesis/communication tasks are hardest, with accuracy capped by messaging and calendar coordination (\emph{Quarterly metrics}, \emph{Colleague meeting}). This preserves difficulty while improving measurability.

\paragraph*{Summary and link to the research question.}
Qualitatively, the new wealth-management suite replaces “can’t reach files” dead-ends with discriminative checks on retrieval, analysis, and delivery. It surfaces the agent’s genuine strengths (CSV parsing, reconciliation) and isolates the true blockers (auth, WebDAV, Rocket.Chat), rather than letting authorisation issues dominate scores. This supports the central research question: an effective suite can be built that reflects a wealth-manager’s assistant role and meaningfully tests an LLM’s ability to act as an employee in a financial firm—\emph{not} by making tasks easier, but by making progress objectively observable and the remaining gaps diagnostic.

\section{Experiment 2 – Task Autonomy Comparison}\label{sec:experiment2}

\begin{figure}[H]
  \centering
  \makebox[\textwidth][c]{%
    \includegraphics[height=0.22\textheight,keepaspectratio]{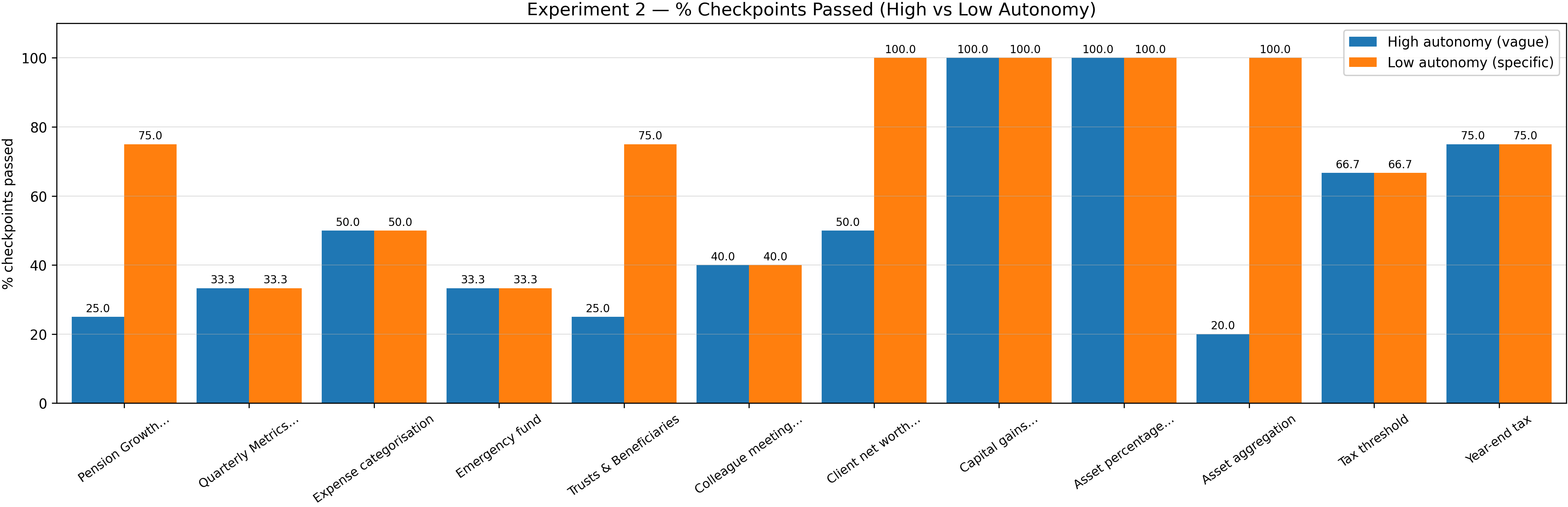}%
  }
  \caption{Experiment 2: \% checkpoints passed — grouped by task (high vs.\ low autonomy).}
  \label{fig:exp2-checkpoints}
\end{figure}

\begin{figure}[H]
  \centering
  \makebox[\textwidth][c]{%
    \includegraphics[height=0.22\textheight,keepaspectratio]{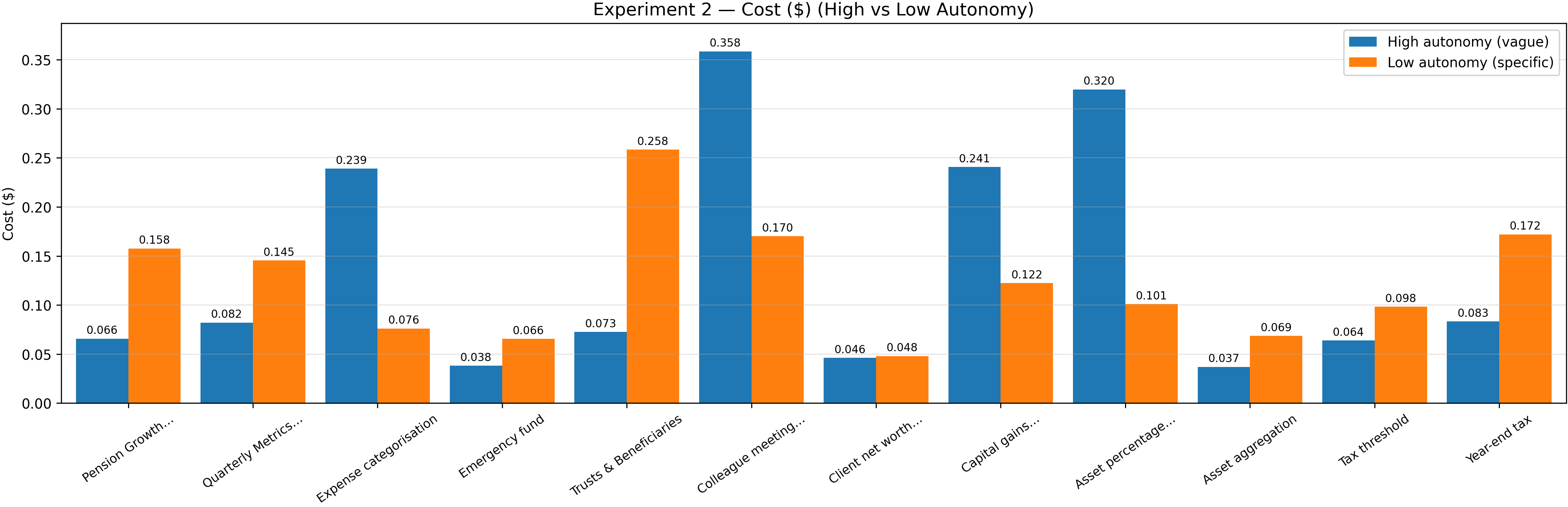}%
  }
  \caption{Experiment 2: Cost (\$) — grouped by task (high vs.\ low autonomy).}
  \label{fig:exp2-cost}
\end{figure}

\begin{figure}[H]
  \centering
  \makebox[\textwidth][c]{%
    \includegraphics[height=0.22\textheight,keepaspectratio]{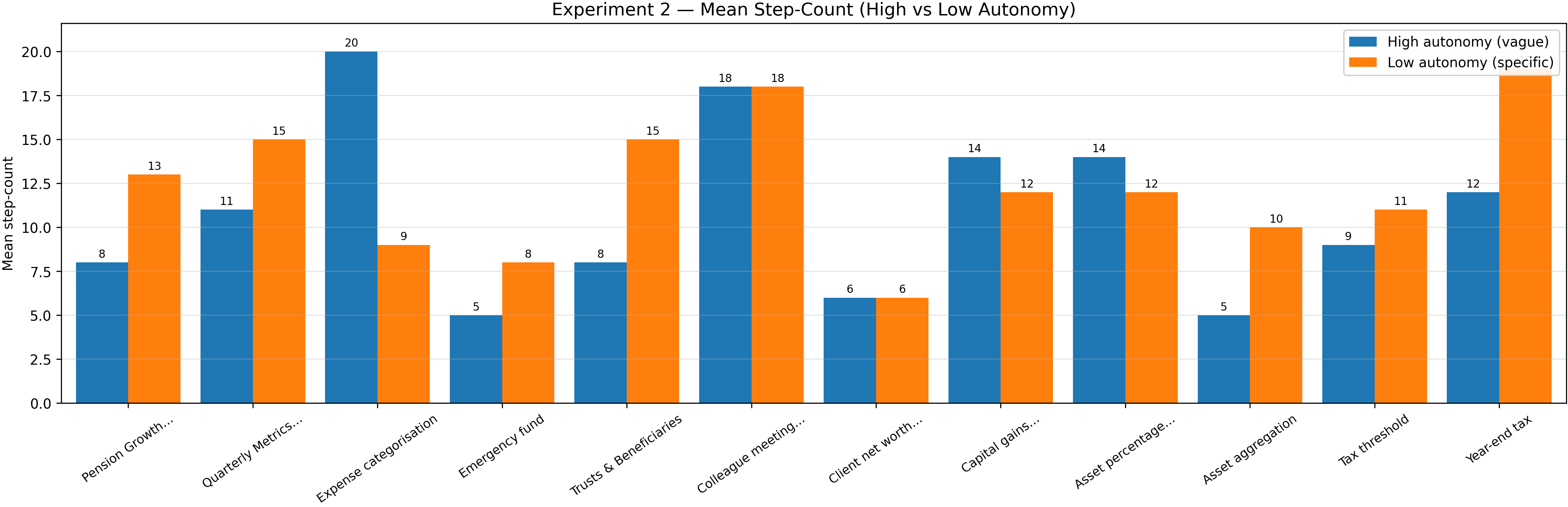}%
  }
  \caption{Experiment 2: Mean step-count — grouped by task (high vs.\ low autonomy).}
  \label{fig:exp2-stepcount}
\end{figure}

\noindent
Figures~\ref{fig:exp2-checkpoints}–\ref{fig:exp2-stepcount} compare the \emph{same 12 tasks} run under two prompt styles: \textbf{high autonomy} (brief/vague) and \textbf{low autonomy} (schema- and path-specific). Bars are paired per task (blue = high autonomy, orange = low autonomy).

\noindent
Checkpoint accuracy improves on most tasks under low autonomy, with large gains on \emph{Pension Growth}, \emph{Trusts \& Beneficiaries}, \emph{Client Net Worth}, and \emph{Asset Aggregation}; several tasks remain flat (\emph{Quarterly Metrics}, \emph{Expense categorisation}, \emph{Emergency fund}, \emph{Tax threshold}, \emph{Year-end tax}).

\noindent
Costs are similar overall; where low autonomy helps the agent avoid failed uploads or retries, it is often slightly \emph{cheaper} despite marginally longer trajectories.

\noindent
Step-counts increase modestly under low autonomy for several tasks (e.g., \emph{Pension Growth}, \emph{Year-end tax}), reflecting extra, directed actions (correct paths, schema checks) rather than thrashing.

\medskip
\noindent

\begin{table}[H]
\centering
\caption{Averages: high autonomy (HA) vs. low autonomy (LA) prompt styles on new tasks. This demonstrates that low autonomy significantly increases checkpoints passed and which are resolved, while not significantly increasing step-count or cost}
\label{tab:vague-vs-specific-avg}
\begin{tabular}{|l|c|c|c|c|}
\hline
\textbf{Prompt Style} & \textbf{Resolved} & \textbf{\% Checkpoints Passed} & \textbf{Mean Step-Count} & \textbf{Cost (\$)} \\
\hline
High Autonomy (HA) & 2/12 & 24/49 = 49.0\% & 10.8 & 0.1373 \\
Low Autonomy (LA)  & 4/12 & 34/49 = 69.4\% & 12.3 & 0.1237 \\
\hline
\end{tabular}
\end{table}

\subsection{Quantitative Analysis}\label{sec:ex2quantiative}

\noindent
Experiment~2 asks: \emph{to what extent does description granularity (low autonomy = schema/path–specific vs.\ high autonomy = brief) affect accuracy and efficiency when the \textbf{same} 12 tasks are run under each style?}
Figures~\ref{fig:exp2-checkpoints}–\ref{fig:exp2-stepcount} and Table~\ref{tab:vague-vs-specific-avg} summarise three signals: checkpoint accuracy, mean step-count, and API cost.

\paragraph*{Overall effect.}
As you can see in Table \ref{tab:vague-vs-specific-avg}, low-autonomy prompts resolve more tasks (4/12 vs.\ 2/12) and raise mean checkpoints passed from 49\% (24/49) to 69\% (34/49), while adding only 1.5 steps per run on average (10.8\,$\rightarrow$\,12.3) and \emph{slightly reducing} mean cost (\$0.1373\,$\rightarrow$\,\$0.1237).%
\footnote{Table~\ref{tab:vague-vs-specific-avg}.} 
The accuracy gain is visible across many bars in Fig.~\ref{fig:exp2-checkpoints}; the efficiency picture (Figs.~\ref{fig:exp2-cost}, \ref{fig:exp2-stepcount}) shows modest trajectory growth but fewer costly retries.

\paragraph*{Difficulty tiers.}
To examine whether granularity helps uniformly, we group tasks by the difficulty design in \S\ref{sec:task-suite-design}:

\begin{itemize}
  \item \textbf{Level~2 (Analytical/Computational)}: \emph{Pension Growth, Expense Categorisation, Trusts \& Beneficiaries, Client Net Worth, Capital Gains, Asset Percentage, Asset Aggregation, Tax Threshold, Year-end Tax}.
  \item \textbf{Level~3 (Synthesis/Communication)}: \emph{Quarterly Metrics}, \emph{Emergency Fund}, \emph{Colleague Meeting Setup}.
\end{itemize}

\noindent
Granularity helps \emph{most} on Level~2 tasks: the average checkpoint pass-rate rises from $\approx$56.9\% (high autonomy) to $\approx$82.4\% (low autonomy).\footnote{Per-task values are the bars in Fig.~\ref{fig:exp2-checkpoints}.} 
In contrast, Level~3 tasks remain flat at $\approx$35.5\% under both styles, indicating that external constraints (authentication, cross-tool delivery) rather than schema inference limit performance. 
This tiered pattern shows the benchmark remains \emph{challenging}: the easier computational wins do not trivialise the suite, while multi-system synthesis still blocks end-to-end success.

\paragraph*{Task-level contrasts }
Several pairs highlight how and when lowered autonomy pays off:
\emph{Pension Growth}, \emph{Trusts \& Beneficiaries}, \emph{Client Net Worth}, and \emph{Asset Aggregation} jump from 25\% to 75\%, 25\% to 75\%, 50\% to 100\%, and 20\% to 100\% respectively when schemas/paths are provided (Fig.~\ref{fig:exp2-checkpoints}). This is consistent with data-reachability being the dominant failure mode. 
Efficiency also improves in \emph{Expense Categorisation} and \emph{Capital Gains}: steps fall (20$\rightarrow$9 and 14$\rightarrow$12) and cost halves (0.239$\rightarrow$0.076 and 0.241$\rightarrow$0.123) because directed access prevents fruitless probing (Figs.~\ref{fig:exp2-cost}, \ref{fig:exp2-stepcount}). 
By contrast, \emph{Quarterly Metrics}, \emph{Emergency Fund}, and \emph{Colleague Meeting} show unchanged accuracy (33–40\%) even under low autonomy; they are bottlenecked by messaging and calendar/issue-tracker actions rather than by file schemas. 
Notably, \emph{Colleague Meeting} keeps the same step-count (18) but halves cost (0.359$\rightarrow$0.170), suggesting fewer failing retries without surpassing checkpoints.

\paragraph*{Answer to the research question.}
Decreased autonomy affects agent accuracy and efficiency, but \emph{selectively}. 
When the task is primarily \emph{analytical on accessible data}, low-autonomy prompts provide large accuracy gains at negligible or favourable cost, likely due to less guess work. 
When the task demands \emph{cross-system synthesis and delivery}, added detail does help overcome difficulties in accessing services in some cases, but this is still a prevalent issue. 
Thus, description granularity is an important control for computational tasks, while solid descriptions do assist in accessing services somewhat, systemic constraints remain the main barrier to realistic assistant work. 
This directly informs tool/benchmark design to use schema-aware prompts and often local input documents to separate “can’t reach the data” from “can’t complete the workflow.”, which helps in evaluating both how the agent performs in accessing services, and its effectiveness in performing the calculations/operations provided to it.

\subsection{Qualitative Analysis}\label{sec:ex2qualitative}
 
\noindent
To avoid repeating task-by-task narratives, we group outcomes by the error patterns that recurred across the same 12 tasks run under \textbf{high autonomy} (brief) and \textbf{low autonomy} (schema/path-specific) prompts. Figures~\ref{fig:exp2-checkpoints}–\ref{fig:exp2-stepcount} and Table~\ref{tab:vague-vs-specific-avg} provide the quantitative context.

\begin{itemize}
    \item \textbf{E1: Data reachability / authentication.}
\emph{Symptom:} agent cannot open required CSVs or APIs (404, 405, login pages), then stalls or fabricates inputs.  
\emph{Where it showed:} strongly under HA in \emph{Pension Growth}, \emph{Trusts \& Beneficiaries}, \emph{Client Net Worth}, \emph{Asset Aggregation}.  
\emph{Effect of LA:} explicit paths and schemas substantially reduced this class of failures, turning HA partials into LA passes (e.g., \emph{Pension Growth} 25\%$\rightarrow$75\%; \emph{Trusts \& Beneficiaries} 25\%$\rightarrow$75\%; \emph{Client Net Worth} 50\%$\rightarrow$100\%; \emph{Asset Aggregation} 20\%$\rightarrow$100\%) but these remained present in both schema.
    \item \textbf{E2: Delivery / upload brittleness.}
\emph{Symptom:} correct local artefacts but failed PUT/WebDAV or misaddressed storage/chat operations.  
\emph{Where it showed:} both styles, notably in \emph{Year-end Tax} and occasionally after HA computation.  
\emph{Effect of LA:} mixed. LA reduces exploratory misfires (lower cost), but cannot commonly overcome server-side constraints; results often remain local-only.
    \item \textbf{E3: Tool integration / messaging.}
\emph{Symptom:} correct calculations but incomplete cross-tool actions (e.g., Rocket.Chat DM not sent; Plane issue not closed).  
\emph{Where it showed:} \emph{Quarterly Metrics}, \emph{Emergency Fund}, \emph{Colleague Meeting}.  
\emph{Effect of LA:} little to none; checkpoints stay flat because the blocker is execution/auth in external tools, not schema discovery.
    \item \textbf{E4: Schema/logic mistakes (with data available).}
\emph{Symptom:} band-capping or double-count errors despite reading the files.  
\emph{Where it showed:} \emph{Tax Threshold} variants (different but offsetting mistakes).  
\emph{Effect of LA:} neutral overall, granularity helps locate fields but does not guarantee correct domain logic without explicit rules.
    \item \textbf{E5: Fabrication fallback.}
\emph{Symptom:} when E1/E3 occur, the model creates fake ledgers or balances to keep the pipeline moving.  
\emph{Where it showed:} HA \emph{Expense Categorisation}, \emph{Asset Aggregation}, \emph{Client Net Worth}.  
\emph{Effect of LA:} sharply reduced because real files are found more reliably; when fabrication persists, content checkpoints fail even if delivery succeeds.

\end{itemize}

\textbf{Cross-style comparison.}
\begin{itemize}
  \item \textbf{Clear LA gains on data-centric L2 tasks:} \emph{Pension Growth}, \emph{Trusts \& Beneficiaries}, \emph{Client Net Worth}, \emph{Asset Aggregation} move from HA partials/failures to LA passes by eliminating E1 (data reachability).
  \item \textbf{Flat on L3 synthesis tasks:} \emph{Quarterly Metrics}, \emph{Emergency Fund}, \emph{Colleague Meeting} remain bottlenecked by E3 (tool execution/DMs/calendars), so granularity does not lift accuracy.
  \item \textbf{Stable parity where computation is already robust:} \emph{Capital Gains}, \emph{Asset Percentage}, \emph{Tax Threshold} show little change in checkpoints; LA primarily trims cost/steps by curbing exploration.
  \end{itemize}

\textbf{Implications for the research question.}
The evidence indicates that \emph{description granularity matters, but selectively}. LA prompts materially improve success on \emph{analytical, data-reachable} workflows by preventing E1 and reducing E5, with small step increases and often lower cost. However, for \emph{multi-system synthesis/delivery} workflows, accuracy is capped by E2–E3 (uploads, messaging, calendars), which prompt detail alone does not fix.  
In terms of the dissertation’s question on autonomy and effectiveness, the qualitative patterns corroborate the quantitative gains: lower autonomy (path specificity and accessibility information) is a high-leverage control for Level-2 tasks, while Level-3 tasks remain challenging and thus valuable for benchmarking end-to-end assistant behaviour.

\chapter{Conclusion \& Future Work}\label{chap:conclusion}         % 7

\section{Concluding Remarks}\label{sec:concluding}
We set out to test whether a general-purpose LLM agent can complete realistic wealth-management assistant workflows inside a reproducible, tool-rich environment. Building on TAC, we added EspoCRM, seeded finance-specific data, paired each task with deterministic evaluators, and introduced a high– vs.\ low-autonomy variant of every task. Rather than making tasks easier, the new suite makes progress \emph{more faithfully measurable}. With the same agent and comparable step-counts, the share of checkpoints credited rose from \(\sim\)15\% to \(\sim\)49\% \emph{because} the evaluators recompute ground truth - for accurate grading, split up checkpoints into more granular steps - for more effective analysis, and awards credit for correctness of answer, even when the agent struggles to access services. Low-autonomy briefs generally improved accuracy further with flat-to-modest cost changes, suggesting that clearer schemas and paths help agents focus on the core computation rather than exploration. Error analyses show most failures cluster around access and delivery (authentication, WebDAV uploads, Rocket.Chat messaging) rather than arithmetic once inputs are available, indicating where systems hardening matters most. Overall, the benchmark yields fair, reproducible scoring: file and numeric checks are fully programmatic; an LLM-as-judge is used only for yes/no content predicates in chat.

Three observations stand out. First, agent competence on finance tasks is limited less by mathematical reasoning than by end-to-end workflow reliability (finding the right files, following naming/location rules, and delivering outputs). Second, autonomy level meaningfully affects outcomes: constraining source, method, and deliverable reduces drift without inflating search depth. Third, strict but tolerant evaluators (schema/format exactness with numeric/CSV normalisation) avoid both false positives and brittle byte-match failures. The work remains limited by a closed-world setup with synthetic, but carefully normalised, data, a small number of clients and tasks relative to industry breadth, and primary evaluation on a single agent configuration. The environment also simplifies some real operational concerns (identity management, heterogeneous file encodings beyond UTF-8, and intermittent service outages).

Several hard problems remain open. Agents struggle with authentication and accessing tools and this is still a major source of failure. Reliable document delivery (especially WebDAV nuances and calendar artefacts) is fragile. Verifying richer free-text communications without over-reliance on LLM judging is another challenge. Finally, maintaining cost efficiency while improving reliability is non-trivial: agents that “try harder” can escalate API spend without proportional accuracy gains.

\section{Possible Future Research Directions}\label{sec:future-research}

We see four concrete future research directions. (i) \textbf{Counterfactual robustness.} This setup provides data in the specified format, and in the correct locations, another direction could be to perturb tasks (header case, benign row reorders, tiny numeric noise, alternate but equivalent schemas, and not in the specified location) this could improve realism further, and provide more valuable insights into agent behaviour. (ii) \textbf{Task and data breadth:} expand beyond assistant tasks toward lightweight advisory previews (e.g., constraints-aware portfolio rebalancing), increase client diversity, and introduce controlled data imperfections (missing values, inconsistent headers) to test robustness. (iii) \textbf{Evaluation science:} refine metrics with cost–accuracy Pareto reporting, add calibrated partial credit for long-form communications (regex + rule checks before any LLM predicate), and open the suite for cross-model comparisons. (iv) \textbf{Agent design:} study memory/persistence and multi-agent collaboration (e.g., planner–executor with handoffs in Plane/Rocket.Chat), and explore light fine-tuning or tool-use adapters targeted at file handling and delivery. Together, these steps would strengthen ecological validity, raise ceiling performance, and keep the benchmark faithful to how real finance teams work while remaining reproducible and economical to run.

%==================== REFERENCES ====================
\bibliographystyle{plain}
\bibliography{mybibfile}

@article{xu2024theagentcompany,
  title        = {TheAgentCompany: Benchmarking LLM Agents on Consequential Real World Tasks},
  author       = {Xu, Frank F. and Song, Yufan and Li, Boxuan and Tang, Yuxuan and Jain, Kritanjali and Bao, Mengxue and Wang, Zora Z. and Zhou, Xuhui and Guo, Zhitong and Cao, Murong and others},
  journal      = {arXiv preprint arXiv:2412.14161},
  year         = {2024}
}

@article{openhands2024,
  title        = {OpenHands: An Open Platform for {AI} Software Developers as Generalist Agents},
  author       = {Wang, Kuan and others},
  journal      = {arXiv preprint arXiv:2407.16741},
  year         = {2024}
}

@inproceedings{drouin2024workarena,
  title        = {WorkArena: Benchmarking Web Agents in Realistic Enterprise Environments},
  author       = {Drouin, Alexandre and Raghu, Dheeraj and Hazra, Rohan and Argal, Achleshwar and others},
  booktitle    = {Proceedings of the 30th ACM SIGKDD Conference on Knowledge Discovery and Data Mining},
  year         = {2024}
}

@article{jimenez2023swebench,
  title        = {{SWE}-bench: Can Language Models Resolve Real-World GitHub Issues?},
  author       = {Jimenez, Nicholas and others},
  journal      = {arXiv preprint arXiv:2310.06770},
  year         = {2023}
}

@article{li2024devbench,
  title        = {DevBench: {A} Comprehensive Benchmark for Software Development},
  author       = {Li, Jia and others},
  journal      = {arXiv preprint arXiv:2403.08604},
  year         = {2024}
}

@article{yoran2024assistantbench,
  title        = {AssistantBench: Evaluating Real-World Personal Assistant Agents},
  author       = {Yoran, Ori and others},
  journal      = {arXiv preprint arXiv:2407.15711},
  year         = {2024}
}

@article{yao2024taubench,
  title        = {$\tau$-bench: A Benchmark for Tool-Agent-User Interaction in Real-World Domains},
  author       = {Yao, Shunyu and Shinn, Noah and Razavi, Pedram and Narasimhan, Karthik},
  journal      = {arXiv preprint arXiv:2406.12045},
  year         = {2024}
}

@article{finben2024,
  title        = {FinBen: A Holistic Financial Benchmark for Large Language Models},
  author       = {Xie, Yutong and others},
  journal      = {OpenReview preprint},
  year         = {2024},
  note         = {\url{https://openreview.net/forum?id=0VJ2uZ8f1A}}
}

@article{bigeard2025fab,
  title        = {Finance Agent Benchmark: Benchmarking {LLMs} on Real-world Financial Research Tasks},
  author       = {Bigeard, B and others},
  journal      = {arXiv preprint arXiv:2508.00828},
  year         = {2025}
}

@article{li2024investorbench,
  title        = {INVESTORBENCH: A Benchmark for Financial Decision-Making Tasks with LLM-based Agent},
  author       = {Li, Enzhi and others},
  journal      = {arXiv preprint arXiv:2412.18174},
  year         = {2024}
}

@article{kapoor2024agentsmatter,
  title        = {AI Agents That Matter},
  author       = {Kapoor, Sayash and Stroebl, Benedikt and Siegel, Zachary S. and Nadgir, Nitya and Narayanan, Arvind},
  journal      = {arXiv preprint arXiv:2407.01502},
  year         = {2024}
}

@misc{gitlab,
  title        = {GitLab: The DevSecOps Platform},
  howpublished = {\url{https://about.gitlab.com/}},
  year         = {2025},
  note         = {Accessed 2025-08-11}
}

@misc{owncloud,
  title        = {ownCloud Documentation},
  howpublished = {\url{https://doc.owncloud.com/}},
  year         = {2025},
  note         = {Accessed 2025-08-11}
}

@misc{rocketchat,
  title        = {Rocket.Chat Documentation},
  howpublished = {\url{https://docs.rocket.chat/}},
  year         = {2025},
  note         = {Accessed 2025-08-11}
}

@misc{plane,
  title        = {Plane: Open-source Project Management},
  howpublished = {\url{https://github.com/makeplane/plane}},
  year         = {2025},
  note         = {Accessed 2025-08-11}
}

@inproceedings{kluyver2016jupyter,
  title        = {Jupyter Notebooks -- a publishing format for reproducible computational workflows},
  author       = {Kluyver, Thomas and Ragan-Kelley, Benjamin and P{\'e}rez, Fernando and Granger, Brian and Bussonnier, Matthias and Frederic, Jonathan and Kelley, Kyle and Hamrick, Jessica and others},
  booktitle    = {Positioning and Power in Academic Publishing: Players, Agents and Agendas},
  year         = {2016},
  publisher    = {IOS Press}
}

@article{openai2023gpt4,
  title        = {GPT-4 Technical Report},
  author       = {OpenAI},
  journal      = {arXiv preprint arXiv:2303.08774},
  year         = {2023}
}

@article{team2023gemini,
  title        = {Gemini: {A} Family of Highly Capable Multimodal Models},
  author       = {Team, Google and others},
  journal      = {arXiv preprint arXiv:2312.11805},
  year         = {2023}
}

@article{dubey2024llama3,
  title        = {The Llama 3 Herd of Models},
  author       = {Dubey, Abhimanyu and others},
  journal      = {arXiv preprint arXiv:2407.21783},
  year         = {2024}
}

@misc{anthropic2024claude35,
  title        = {Introducing Claude 3.5 Sonnet},
  author       = {{Anthropic}},
  howpublished = {\url{https://www.anthropic.com/news/claude-3-5-sonnet}},
  year         = {2024},
  note         = {Accessed 2025-08-11}
}

@article{wang2024llm_agent_survey,
  title        = {A Survey on Large Language Model based Autonomous Agents},
  author       = {Wang, Yutao and others},
  journal      = {Frontiers of Computer Science},
  year         = {2024},
  doi          = {10.1007/s11704-024-36236-3},
  note         = {\url{https://link.springer.com/article/10.1007/s11704-024-36236-3}}
}

@book{russell2020aima,
  title     = {Artificial Intelligence: A Modern Approach},
  author    = {Russell, Stuart and Norvig, Peter},
  edition   = {4th},
  publisher = {Pearson},
  year      = {2020}
}

@article{yao2023react,
  title   = {ReAct: Synergizing Reasoning and Acting in Language Models},
  author  = {Yao, Shunyu and Zhao, Jeffrey and Yu, Dian and others},
  journal = {arXiv preprint arXiv:2210.03629},
  year    = {2023}
}

@article{schick2023toolformer,
  title   = {Toolformer: Language Models Can Teach Themselves to Use Tools},
  author  = {Schick, Timo and Dwivedi-Yu, Jane and others},
  journal = {arXiv preprint arXiv:2302.04761},
  year    = {2023}
}

@article{shinn2023reflexion,
  title   = {Reflexion: Language Agents with Verbal Reinforcement Learning},
  author  = {Shinn, Noah and others},
  journal  = {arXiv preprint arXiv:2303.11366},
  year    = {2023}
}

@article{wang2023voyager,
  title   = {Voyager: An Open-Ended Embodied Agent with Large Language Models},
  author  = {Wang, Guanzhi and others},
  journal = {arXiv preprint arXiv:2305.16291},
  year    = {2023}
}

@misc{espocrm,
  title        = {EspoCRM Documentation},
  howpublished = {\url{https://www.espocrm.com/documentation/}},
  year         = {2025},
  note         = {Accessed 2025-08-11}
}

@misc{mckinsey2023genaiBanking,
  title        = {Capturing the full value of generative AI in banking},
  author       = {{McKinsey \& Company}},
  year         = {2023},
  howpublished = {\url{https://www.mckinsey.com/industries/financial-services/our-insights/generative-ai-in-banking-a-200-billion-opportunity}},
  note         = {Accessed 2025-08-21}
}

%==================== APPENDICES ====================
\chapter{Appendix}

\section{Task Catalogue (Overview)}\label{app:task-catalogue}

\small
\renewcommand{\arraystretch}{1.15}
\setlength{\tabcolsep}{4pt}

\begin{longtable}{@{}p{0.20\textwidth} p{0.26\textwidth} p{0.26\textwidth} p{0.26\textwidth}@{}}
\caption{Task overview: purpose, inputs, outputs, and tools}\label{tab:task-overview}\\
\toprule
\textbf{Task (Level)} & \textbf{Purpose} & \textbf{Inputs} & \textbf{Outputs / Tools} \\
\midrule
\endfirsthead
\toprule
\textbf{Task (Level)} & \textbf{Purpose} & \textbf{Inputs} & \textbf{Outputs / Tools} \\
\midrule
\endhead
\midrule \multicolumn{4}{r}{\itshape continued on next page} \\
\endfoot
\bottomrule
\endlastfoot

\textbf{Net Worth Snapshot (L1)} &
Provide a snapshot of a client’s net worth by summarising assets and liabilities. &
CSV files from \emph{OwnCloud} (\texttt{assets.csv}, \texttt{liabilities.csv}) &
TXT file in \emph{OwnCloud} (\texttt{net\_worth\_snapshot.txt}); \emph{OwnCloud} \\
\addlinespace

\textbf{Asset Aggregation (L2)} &
Aggregate the client’s cash, investment, and property balances into one total-assets figure. &
CSVs from \emph{OwnCloud} (\texttt{accounts.csv}, \texttt{investments.csv}, \texttt{property\_holdings.csv}) &
TXT file in \emph{OwnCloud} (\texttt{total\_assets.txt}); \emph{OwnCloud} \\
\addlinespace

\textbf{Year-End Tax Summary (L2)} &
Summarise taxable income data for client tax reporting. &
CSVs from \emph{OwnCloud} (\texttt{income.csv}, \texttt{gains.csv}, \texttt{dividends.csv}) &
CSV in \emph{OwnCloud} (\texttt{YearEnd\_Tax\_Report\_2024.csv}); \emph{OwnCloud} \\
\addlinespace

\textbf{Trusts \& Beneficiaries (L2)} &
Identify discrepancies in trust beneficiary information. &
CSVs from \emph{EspoCRM} (\texttt{beneficiaries.csv}, \texttt{trusts.csv}); deed from \emph{OwnCloud} (\texttt{trust\_deed.csv}) &
CSV in \emph{OwnCloud} (\texttt{\textless trust\_id\textgreater\_beneficiary\_discrepancies.csv}); \emph{EspoCRM}, \emph{OwnCloud} \\
\addlinespace

\textbf{Tax Threshold Calculation (L2)} &
Calculate the client’s total tax liability based on their estate and assets. &
CSV/TXT from \emph{OwnCloud} (\texttt{estate\_summary.csv}, \texttt{assets.csv}, \texttt{rates\_and\_thresholds.txt}) &
One-line TXT in \emph{OwnCloud} (\texttt{tax\_due.txt}); \emph{OwnCloud} \\
\addlinespace

\textbf{Pension Projection (L2)} &
Project client’s pension portfolio growth until retirement. &
Data from \emph{EspoCRM} (\texttt{pension\_contributions.csv}, \texttt{growth\_rates.txt}) &
CSV in \emph{OwnCloud} (\texttt{pension\_projection.csv}); \emph{EspoCRM}, \emph{OwnCloud} \\
\addlinespace

\textbf{Expense Categorisation (L2)} &
Categorise client spending to identify potential savings opportunities. &
CSV from \emph{EspoCRM} (\texttt{transactions.csv}) &
CSV in \emph{OwnCloud} (\texttt{transactions\_categorised.csv}); \emph{EspoCRM}, \emph{OwnCloud} \\
\addlinespace

\textbf{Capital Gains Computation (L2)} &
Calculate gains or losses from a client’s trading activity using FIFO accounting. &
CSV from \emph{OwnCloud} (\texttt{trades.csv}) &
CSV in \emph{OwnCloud} (\texttt{capital\_gains.csv}); \emph{OwnCloud} \\
\addlinespace

\textbf{Portfolio Asset Allocation (L2)} &
Analyse and summarise the client’s investment portfolio by asset class. &
CSV from \emph{OwnCloud} (\texttt{portfolio.csv}) &
CSV in \emph{OwnCloud} (\texttt{asset\_percentages.csv}); \emph{OwnCloud} \\
\addlinespace

\textbf{Emergency Fund (L3)} &
Determine if a colleague’s savings cover at least six months of expenses. &
CSVs from \emph{OwnCloud} (\texttt{expenses\_summary.csv}, \texttt{savings\_balance.csv}) &
Structured message via \emph{Rocket.Chat}; \emph{OwnCloud}, \emph{Rocket.Chat} \\
\addlinespace

\textbf{Meeting Scheduling Coordination (L3)} &
Schedule client meetings by aligning client and advisor calendars. &
Calendar PDF from \emph{EspoCRM}; calendar via \emph{Rocket.Chat} &
\emph{Rocket.Chat} confirmation; ICS in \emph{OwnCloud} (\texttt{ClientHoldingReview.ics}); \emph{EspoCRM}, \emph{Rocket.Chat}, \emph{OwnCloud} \\
\addlinespace

\textbf{Quarterly Metrics Analysis \& Communication (L3)} &
Analyse quarterly performance metrics and communicate insights to stakeholders. &
CSV from \emph{OwnCloud} (\texttt{Q2\_metrics.csv}); stakeholder list &
Structured messages via \emph{Rocket.Chat}; task completion confirmation in \emph{Plane}; \emph{Plane}, \emph{OwnCloud}, \emph{Rocket.Chat} \\
\end{longtable}

\section{Checkpoint Descriptions (Per Task)}\label{app:checkpoint-descriptions}
% Derived from the original methodology checkpoint section.  [oai_citation:1‡3 Methodology and System Design.docx](file-service://file-51ch6Wt2mMh1gTWk3adpXJ)
\small

\subsection*{Net Worth Snapshot (4 points)}
\begin{enumerate}\item (1 pt) Navigate to \texttt{/Finance\_Documents/Clients/MikeLazaridis/} and open \texttt{assets.csv} and \texttt{liabilities.csv}.
\item (1 pt) Correctly sum total assets and liabilities.
\item (2 pts) Write \texttt{net\_worth\_snapshot.txt} containing exactly \texttt{Net worth: £<value>} with the correct figure and no additional text.
\end{enumerate}

\subsection*{Year-End Tax Summary (4 points)}
\begin{enumerate}\item (1 pt) Access \texttt{income.csv}, \texttt{gains.csv}, and \texttt{dividends.csv} in \texttt{/Finance\_Documents/Clients/RichardCurtis/}.
\item (1 pt) Calculate sums for each income category accurately.
\item (1 pt) Populate the \texttt{YearEnd\_Tax\_Report\_2024.csv} template correctly.
\item (1 pt) Ensure the final CSV adheres to the provided template’s schema and formatting.
\end{enumerate}

\subsection*{Trusts and Beneficiaries (4 points)}
\begin{enumerate}\item (1 pt) Retrieve \texttt{beneficiaries.csv} and \texttt{trusts.csv} from EspoCRM, plus \texttt{trust\_deed.csv} from OwnCloud (or locally).
\item (2 pts) Generate \texttt{<trust\_id>\_beneficiary\_discrepancies.csv} with columns \texttt{Beneficiary\_Name}, \texttt{CRM\_Percent}, \texttt{Deed\_Percent} and correct values.
\item (1 pt) Save the output in the correct client directory with filename based on trust ID.
\end{enumerate}

\subsection*{Tax Threshold Calculation (3 points)}
\begin{enumerate}\item (1 pt) Open \texttt{estate\_summary.csv}, \texttt{assets.csv}, and \texttt{rates\_and\_thresholds.txt}.
\item (1 pt) Create \texttt{tax\_due.txt} with \texttt{Tax due: £<amount>} using correct calculations.
\item (1 pt) Place \texttt{tax\_due.txt} in the designated folder.
\end{enumerate}

\subsection*{Pension Growth Contributions (4 points)}
\begin{enumerate}\item (1 pt) Access \texttt{pension\_contributions.csv} and \texttt{growth\_rates.txt} from EspoCRM.
\item (1 pt) Generate \texttt{pension\_projection.csv} with columns \texttt{Year} and \texttt{Portfolio value (£)}.
\item (2 pts) Ensure each year’s projection and the final value are mathematically accurate.
\end{enumerate}

\subsection*{Expense Categorisation (4 points)}
\begin{enumerate}\item (1 pt) Retrieve \texttt{transactions.csv} from EspoCRM.
\item (1 pt) Upload the enriched dataset to OwnCloud at \texttt{/Finance\_Documents/Clients/MikeReeves/}.
\item (1 pt) Create \texttt{transactions\_categorised.csv} including the new \texttt{Category} column.
\item (1 pt) Correctly assign each transaction as Fixed, Discretionary, or Investment.
\end{enumerate}

\subsection*{Capital Gains Computation (4 points)}
\begin{enumerate}\item (1 pt) Open \texttt{trades.csv} in OwnCloud.
\item (1 pt) Compute per-trade gain/loss rows in \texttt{capital\_gains.csv}.
\item (1 pt) Save \texttt{capital\_gains.csv} in the correct client folder.
\item (1 pt) Append a final row with the total gain or loss.
\end{enumerate}

\subsection*{Portfolio Asset Allocation (3 points)}
\begin{enumerate}\item (1 pt) Access \texttt{portfolio.csv} from OwnCloud.
\item (1 pt) Calculate each asset type’s percentage of the total.
\item (1 pt) Save \texttt{asset\_percentages.csv} with correct columns and values.
\end{enumerate}

\subsection*{Emergency Fund (3 points)}
\begin{enumerate}\item (1 pt) Retrieve \texttt{expenses\_summary.csv} and \texttt{savings\_balance.csv}.
\item (1 pt) Calculate coverage ratio and determine PASS/FAIL (\(\geq\)6 months).
\item (1 pt) Send a RocketChat DM formatted as \texttt{Status: PASS/FAIL, Coverage ratio: <number>}.
\end{enumerate}

\subsection*{Meeting Scheduling Coordination (5 points)}
\begin{enumerate}\item (1 pt) Request and receive the advisor’s calendar in RocketChat.
\item (1 pt) Download the client’s PDF calendar from EspoCRM.
\item (2 pts) Identify the earliest 30-minute slot within 09:00–17:00.
\item (1 pt) Confirm via RocketChat and save \texttt{ClientHoldingReview.ics} in OwnCloud.
\end{enumerate}

\subsection*{Quarterly Metrics Analysis (6 points)}
\begin{enumerate}\item (1 pt) Detect the “Analyze Q2 Metrics” issue in Plane.
\item (1 pt) Download \texttt{Q2\_metrics.csv} and compute YoY growth rates.
\item (1 pt) Draft a two-sentence insight in \texttt{<Metric: Value>} format.
\item (2 pts) DM stakeholders listed in \texttt{stakeholder\_list.csv} with the insight.
\item (1 pt) Mark the Plane task as done.
\end{enumerate}

\subsection*{Asset Aggregation (5 points)}
\begin{enumerate}\item (1 pt) Navigate to \texttt{/Finance\_Documents/Clients/JohnSmith/} on OwnCloud and open \texttt{accounts.csv}, \texttt{investments.csv}, and \texttt{property\_holdings.csv}.
\item (1 pt) Create \texttt{total\_assets.txt} in the same directory.
\item (3 pts) Ensure \texttt{total\_assets.txt} contains exactly one line \texttt{Total assets: £<sum>} where \texttt{<sum>} aggregates all balances precisely.
\end{enumerate}

\section{Data Generation: File Families and Schemas}\label{app:data-generation-table}
% Converted from the methodology data-generation table.  [oai_citation:2‡3 Methodology and System Design.docx](file-service://file-51ch6Wt2mMh1gTWk3adpXJ)
\small
\renewcommand{\arraystretch}{1.15}
\setlength{\tabcolsep}{4pt}
\begin{longtable}{|p{0.22\textwidth}|p{0.17\textwidth}|p{0.28\textwidth}|p{0.27\textwidth}|}
\caption{Task‐family, primary store, object types, and high-level schemas}\label{tab:app-data-schemas}\\
\hline
\textbf{Task family} & \textbf{Primary store} & \textbf{File / object type} & \textbf{High-level schema} \\
\hline
\endfirsthead
\hline
\textbf{Task family} & \textbf{Primary store} & \textbf{File / object type} & \textbf{High-level schema} \\
\hline
\endhead
\hline \multicolumn{4}{r}{\itshape continued on next page} \\
\endfoot
\hline
\endlastfoot

Asset / liability snapshots & OwnCloud
& \texttt{accounts.csv}, \texttt{investments.csv}, \texttt{property\_holdings.csv}
& CSV, 2 columns: Label, Value (£) \\ \hline

Net-worth \& tax tasks & OwnCloud
& \texttt{assets.csv}, \texttt{liabilities.csv}, \texttt{estate\_summary.csv}
& Same 2-column CSV; values may include £ and thousands separators \\ \hline

Portfolio composition & OwnCloud
& \texttt{portfolio.csv}
& CSV, 3 columns: Asset Name, Asset Type, Value (£) \\ \hline

Pension projection & EspoCRM
& \texttt{pension\_contributions.csv}, \texttt{growth\_rates.txt}
& Contributions: single numeric column; Growth rates: Year / Rate \% \\ \hline

Capital-gains history & OwnCloud
& \texttt{trades.csv}
& CSV, 6 columns: Stock, Buy\_Date, Sell\_Date, Quantity, Buy\_Price, Sell\_Price \\ \hline

Spending classification & EspoCRM
& \texttt{transactions.csv}
& CSV, 3 columns: Date, Amount, Description \\ \hline

Emergency-fund check & OwnCloud
& \texttt{expenses\_summary.csv}, \texttt{savings\_balance.csv}
& Both: two-column tables Month/Account vs Value (£) \\ \hline

Trusts and Beneficiaries & EspoCRM + OwnCloud
& \texttt{trusts.csv}, \texttt{beneficiaries.csv}, \texttt{trust\_deed.csv}
& Tidy CSVs with IDs, names, percent fields (with \%) \\ \hline

Year-end tax report & OwnCloud
& \texttt{income.csv}, \texttt{gains.csv}, \texttt{dividends.csv}, template
& Category CSVs: Source / Amount; Template: fixed CSV skeleton \\ \hline

Quarterly metrics & OwnCloud
& \texttt{Q2\_metrics.csv}
& CSV, 3 columns: Metric, Current, Prior \\ \hline

Stakeholder IDs & OwnCloud
& \texttt{stakeholder\_list.csv}
& CSV, 2 columns: Name, RocketChat\_ID \\ \hline

Colleague Meeting Setup & EspoCRM + RocketChat
& Calendar PDF + RocketChat calendar message
& Calendar of availability for a specified week \\ \hline
\end{longtable}

\end{document}